\documentclass[]{lab}

\usepackage[utf8]{inputenc}
\usepackage[T1]{fontenc}
\usepackage{geometry}
\usepackage{amsmath,amssymb}  %
\usepackage{charter}

\usepackage[toc,page,header]{appendix}

\usepackage{minitoc}
\usepackage{cleveref} 
\usepackage{subcaption}
\usepackage{booktabs}
\usepackage{graphicx}
\usepackage{pgfplots}
\usepackage{pgfplotstable}
\usepackage{xcolor}
\usepackage{CJKutf8}
\usetikzlibrary{patterns}
\usepackage{float}
\usepackage{caption}
\usepackage{bm}
\usepackage{makecell} 
\usepackage{tabularx}
\usepackage{colortbl}

\usepackage{soul} %
\usepackage{algorithm}      %
\usepackage{algpseudocode}  %
\usepackage{parskip}

\definecolor{cvprblue}{rgb}{0.21,0.49,0.74}
\definecolor{fallbackgreen}{rgb}{130, 180, 102}
\definecolor{stopred}{rgb}{251, 225, 224}

\ifdefined\final
\usepackage[disable]{todonotes}
\else
\usepackage[textsize=tiny]{todonotes}
\fi

\newcommand{\previous}{\texttt{Vidu S1}\xspace}
\newcommand{\ours}{\texttt{Vidu S2}\xspace}
\newcommand{\oursa}{\texttt{Vidu S2-Avatar}\xspace}
\newcommand{\ourse}{\texttt{Vidu S2-Editing}\xspace}
\DeclareTextFontCommand{\tabletextbf}{\bfseries}
\newcommand{\tablebodyfont}{%
  \normalfont\normalsize
  \let\textbf\tabletextbf
  \renewcommand{\ours}{Vidu S2\xspace}%
  \renewcommand{\oursa}{Vidu S2-Avatar\xspace}%
  \renewcommand{\ourse}{Vidu S2-Editing\xspace}%
}
\AtBeginEnvironment{table}{\tablebodyfont}
\AtBeginEnvironment{table*}{\tablebodyfont}
\DeclareRobustCommand{\tableshading}[2]{%
  \begingroup\setlength{\fboxsep}{1pt}\colorbox{#1}{#2}\endgroup}

\usepackage{natbib}
\usepackage{latexsym}

\usepackage{url}
\usepackage{amssymb}
\usepackage[utf8]{inputenc}
\usepackage{microtype}
\usepackage{booktabs}
\usepackage{pifont} 
\usepackage{multirow}
\usepackage{makecell}
\usepackage{paralist}
\usepackage{xspace}
\usepackage{color}
\usepackage{xcolor}
\usepackage{colortbl}
\usepackage{adjustbox}
\usepackage{hyperref} 
\usepackage[edges]{forest}
\usepackage{tikz} 
\usepackage{caption}
\usepackage{amsfonts}

\hypersetup{
    colorlinks,
    linkcolor={blue!80!black},
    citecolor={blue!80!black},
}
\tikzset{
    root/.style =             {align=center, text width=1cm, rounded corners=3pt, line width=0.3mm, fill=gray!10, draw=gray!80, font=\small},
    demographic/.style =         {align=center, text width=1.8cm, rounded corners=3pt, line width=0.3mm, fill=blue!10, draw=blue!80, font=\footnotesize},
    demographic_work/.style =    {align=center, text width=10cm, rounded corners=3pt, line width=0.3mm, fill=blue!10, draw=blue!0, font=\footnotesize},
    character/.style =         {align=center, text width=1.8cm, rounded corners=3pt, line width=0.3mm, fill=red!10, draw=red!80, font=\footnotesize},
    character_work/.style =    {align=center, text width=10cm, rounded corners=3pt, line width=0.3mm, fill=red!10, draw=red!0, font=\footnotesize},
    personalization/.style =           {align=center, text width=1.8cm, rounded corners=3pt, line width=0.3mm, fill=cyan!10, draw=cyan!80, font=\footnotesize},
    personalization_work/.style =      {align=center, text width=10cm, rounded corners=3pt, line width=0.3mm, fill=cyan!10, draw=cyan!0, font=\footnotesize},
    risk/.style =         {align=center, text width=1.8cm, rounded corners=3pt, line width=0.3mm, fill=orange!10, draw=orange!80, font=\footnotesize},
    risk_work/.style =    {align=center, text width=10cm, rounded corners=3pt, line width=0.3mm, fill=orange!10, draw=orange!0, font=\footnotesize},
}

\newcommand{\reference}[1]{\cref{#1}}

\usepackage{CJK}

\renewcommand{\thefootnote}{}

\newtcolorbox{promptbox}[1][]{
  enhanced,
  breakable,
  colback=promptboxlightgray,
  colframe=promptboxblue!30,
  arc=8pt,
  boxrule=0.5pt,
  left=12pt,
  right=12pt,
  top=8pt,
  bottom=8pt,
  fonttitle=\bfseries,
  fontupper=\linespread{1.2}\selectfont,
  title=#1
}

\title{Vidu S2: Real-Time Interactive, Editable, and Spatial Video Generation}

\author{{\fontsize{10.5}{11}\selectfont Jintao Zhang$^{*\dagger}$, Kai Jiang$^{*}$, Jintao Chen$^{*}$, Xu Wang$^{*}$, Deyuan Liu$^{*}$, Jungang Li$^{*}$, Dechuang Chen$^{*}$, Ming Lin$^{*}$, Jingjiang Zhou, Haopeng Jin, Qi Jia, Xiaohang Wang, Yaole Wang, Zhanqiang Zhang, Ran Li, Zhengkun Huang, Shuyue Xiong, Yuji Wang, Zikun Dai, Hui He, Yang Luo, Mang Ning, Weiqi Feng, Chengyang Ye, Xinyue Lin, Min Zhao, Hongzhou Zhu, Hengkai Tan, Zeyuan Wang, Chendong Xiang, Kaiwen Zheng, Zhijie Deng$^{\ddagger}$, Fan Bao$^{\ddagger}$, Jianfei Chen$^{\ddagger}$, Jun Zhu$^{\ddagger}$}}

\affiliation{Tsinghua University, Shengshu Technology\\\url{https://vidu.com/vidu-stream}}

\vspace{-10pt}
\abstract{
We present \ours, which comprises \oursa, a real-time interactive digital-character model, and \ourse, a real-time video editing model. Moreover, we explore the feasibility of real-time spatial video generation for both \oursa and \ourse. Compared with \previous, \oursa supports real-time 720p video generation, generation with dynamic references that can be updated at any moment, and stronger instruction following, such as dancing. \ourse supports editing a video stream in real time, including style transfer, virtual try-on, character replacement, and background replacement. Experiments show that \ours outperforms all baselines. A playable online demo is available at \url{https://vidu.com/vidu-stream}.
}

\begin{document}

\maketitle

\renewcommand{\thefootnote}{}
\footnotetext{*Core contributors, co-first authorship. $^\dagger$Project lead. $^\ddagger$Advisors.}

\renewcommand{\thefootnote}{\arabic{footnote}}

\begin{figure}[H]
    \vspace{-20pt}
    \centering
    \includegraphics[width=\linewidth]{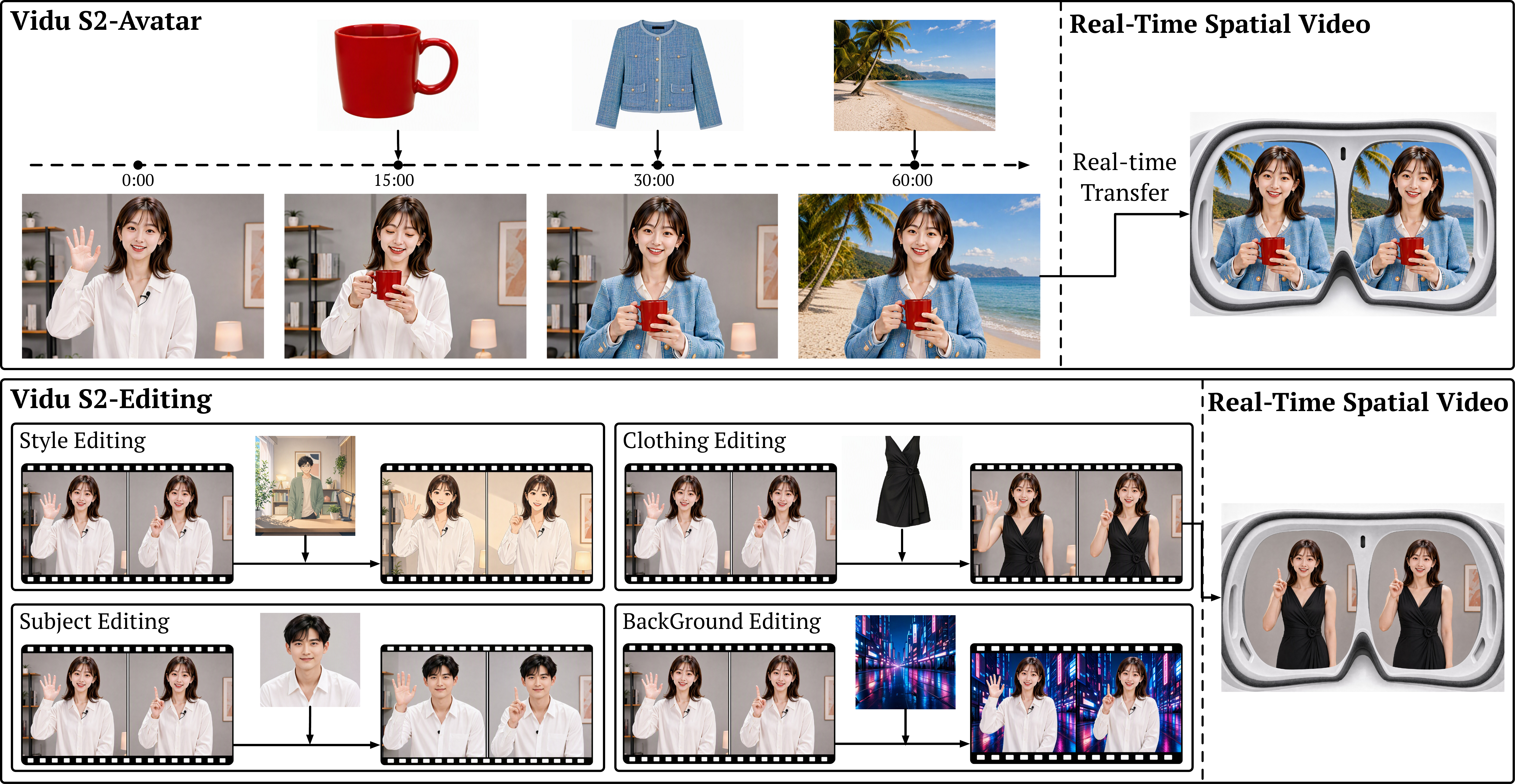}
    \vspace{-1.5em}
    \caption{Overview of \ours.}
    \label{fig:teaser}
\end{figure}

\section{Introduction}

\paragraph{Demand for Real-time Interactive Video Generation}
Recent video generation models, such as Sora, Veo, Wan, and Seedance~\citep{videoworldsimulators2024,veo,wan2025wan,seedance2026seedance}, have shown strong ability in generating high-quality videos. However, most of them still follow an offline, one-shot generation paradigm: a user enters a prompt, waits for minutes or even tens of minutes, and receives the complete video only after generation finishes. 
During this process, the user can only passively wait to receive information and cannot actively initiate any interaction.
This limitation comes from the offline diffusion paradigm, where the model denoises the entire video synchronously over many steps and only produces an entire clean video at the end. Such a paradigm works well for offline content creation, but human visual entertainment is not limited to pre-generated videos. People also enjoy face-to-face communication, live streaming, games, talking with someone, and other interactive visual experiences, where content must respond immediately to the user. From a demand perspective, suppose that each user has an average demand of $\alpha \in [0,1]$ for real-time interactive visual content, e.g., $\alpha = 0.5$. Then the total demand scales with $\alpha \times N$, where $N$ is the number of users. In contrast, suppose that each user has an average demand of $\beta \in [0,1]$ for offline-generated visual content, e.g., $\beta = 0.5$. Since offline-generated videos can be replayed and shared, their generation demand scales more like $\beta \times N / m$, where $m$ is the average number of views per generated video. If we assume $\alpha \approx \beta$ and $m > 100$, then the demand for real-time interactive video generation is much greater than that for offline-generated videos.

\begin{center}
\underline{\textbf{Real-time interactive video generation is one of the most important future directions for industry.}}
\end{center}
\vspace{-5pt}

\vspace{-1pt}
\paragraph{Vidu S1.}
Following this direction, we released \previous~\citep{zhang2026vidu}, a real-time interactive video generation model for continuous user interaction. Users can control video generation content at any moment through voice instructions, instead of fixing all controls before generation starts. \previous supports infinite-length real-time video generation without blurring, drift, or visual distortion. Built with TurboDiffusion~\citep{zhang2025turbodiffusion} and TurboServe~\citep{jiang2026turboserve}, \previous outputs 540p real-time videos at up to 42 FPS on regular consumer GPUs. Users can upload custom images of real people, anime, and pets, and choose different voice tones for personalized experiences. Its scope, however, is largely confined to talking-head-centric digital characters: the resolution is limited to 540p, the reference that defines the character is fixed once a stream has started, large body motion such as dancing is difficult to follow, and editing an incoming video stream is not supported.

\vspace{-3pt}
\paragraph{Vidu S2-Avatar.}
\oursa is a real-time interactive digital-character model that improves on \previous in four directions. (1) \textit{Self-Replay Forcing (SRF).} A stream is generated segment by segment, so errors pass from one segment to the next and eventually cause drift or collapse. Self-Forcing~\citep{huang2026self} gets the setup right by conditioning each segment on chunks that the model generated itself, which matches what the model sees at inference time. Two problems remain: the self-generated history is fed in clean rather than noised, and it is detached from the computation graph, so no gradient flows through it. Both limit data efficiency and training quality. 
We therefore introduce Self-Replay Forcing (SRF).
After the student performs a long autoregressive rollout, we take the entire student-generated trajectory, independently re-noise all of its segments following Diffusion Forcing \cite{DF}, and replay the full trajectory in a single gradient-enabled causal pass. 
The original rollout and its KV caches are detached before replay, so gradients do not backpropagate through the rollout itself.
Instead, all replayed segments remain connected within the same computation graph, allowing the loss of a later segment to propagate to preceding segments during the replay pass.
(2) \textit{720p Resolution.} \oursa raises real-time generation from 540p to 720p while keeping 25\textasciitilde 42 FPS. A lightweight Refiner adds a single step to lift the resolution, so the backbone can keep running fast at low resolution. We also select training videos by measured clarity instead of nominal resolution, because heavily compressed footage looks blurry even at 1080p. (3) \textit{Stronger instruction following.} \oursa follows a much wider range of instructions, including large body motion such as dancing. We add solo dance videos and 2D/3D animation to the training data, keep videos whose camera moves by stabilizing their backgrounds rather than discarding them, and caption each clip in the order that events happen. Reinforcement learning from human preference then further improves motion naturalness, expressiveness, and instruction adherence. (4) \textit{Reference interaction at any moment.} Users can give the model a new reference image at any point in a stream, for example, an object to pick up, a piece of clothing to put on, or a background to move to. Beyond training on data built for reference-conditioned generation, we further build a VLM agentic system to make the generation more reliable. 

\vspace{-3pt}
\paragraph{Vidu S2-Editing.}
\ourse edits a video stream in real time. It can (1) repaint the whole video in a new visual style, such as turning a real-world video into an anime look, (2) change the clothes the person is wearing, (3) replace the person with a different character, and (4) replace the background behind the person. Each edit follows a text instruction and an optional reference image that shows the desired appearance. The key design is frame-aligned attention: every target frame reads only the source frame at the same time step, so the edited video keeps exactly the same motion and timing as the input, while the reference image stays visible to all frames so that the new appearance carries through the whole video. During streaming, each source frame is consumed together with the target frame it produces and is not kept in the cache.

\vspace{-3pt}
\paragraph{Real-Time Spatial Video Generation.}
We further explore the feasibility of real-time spatial video generation and
editing with \oursa and \ourse. (1) \textit{Vidu S2-Avatar.} The stream
generated by \oursa can be converted into synchronized left- and right-eye
views within the streaming pipeline. This enables real-time interaction with
generated characters in spatial-video form. (2) \textit{Vidu S2-Editing.}
The editing pipeline supports two input formats. For monocular input, \ourse
can edit the stream first and then apply the same conversion. For stereoscopic
input, it can jointly edit paired views and
then split the output back into left- and right-eye views. Across both editing
settings, users can perform style transfer, virtual try-on, character
replacement, and background replacement. The generated or edited views can
then be streamed to VR head-mounted displays, where users can experience the
results as immersive, continuously updated spatial video.

\vspace{2pt}
\begin{center}
    \underline{\textbf{Vidu S is dedicated to creating the ultimate interactive visual experience for humans.}}
\end{center}

\vspace{-2pt}
\paragraph{Contribution.}
We summarize our contributions as follows.

\begin{enumerate}
\item We introduce \oursa, a real-time interactive digital-character model that supports 720p real-time generation, dynamic references that can be updated at any moment during a stream, and stronger instruction following over a wider range of body motion, such as dancing.

\item We introduce \ourse, a real-time video editing model that edits an incoming video stream on the fly, covering style transfer, virtual try-on, character replacement, and background replacement.

\item We explore real-time spatial video generation and editing for VR head-mounted displays. Our streaming framework can convert avatar-generated or edited monocular streams into synchronized left- and right-eye views, or directly edit existing spatial video.

\item We build an efficient inference and serving stack that makes these models practical on low-cost GPUs. It uses SageAttention, SpargeAttention, and Sparse-Linear Attention, low-bit GEMM, kernel fusion and launch optimization, and multi-GPU parallelism that lets the VAE encoder, backbone, Refiner, and VAE decoder share GPUs along a common timeline.

\item Experiments show that \ours outperforms all baselines while fully meeting real-time inference requirements.

\end{enumerate}

\section{Vidu S2-Avatar}

\begin{figure}[H]
    \centering
    \includegraphics[width=\linewidth]{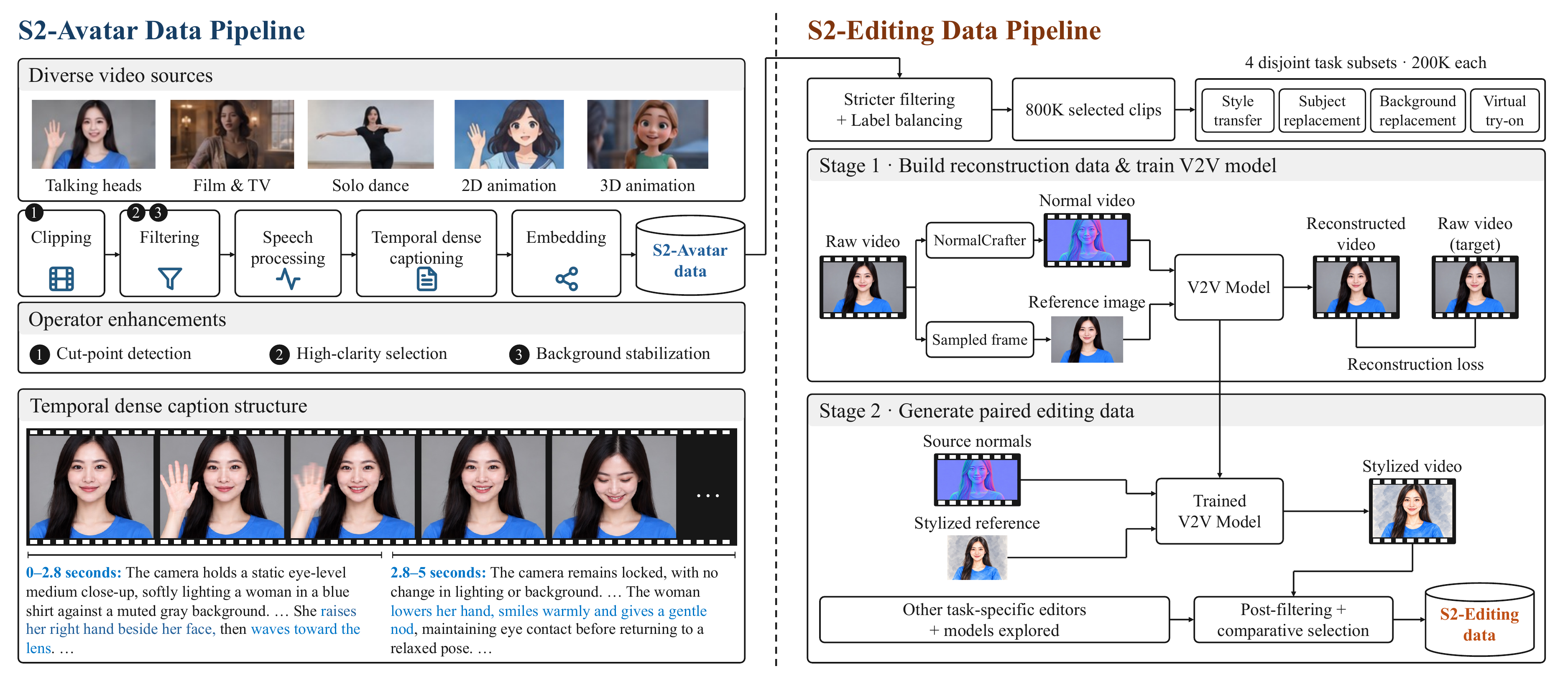}
    \caption{Overview of the data preparation pipelines for \oursa and \ourse. Left: diverse video sources are processed through clipping, filtering, speech processing, temporal dense captioning, and embedding, with enhanced operators for cut-point detection, high-clarity selection, and background stabilization. Right: filtered Avatar data are further curated for four editing tasks; style-transfer pairs are generated through reconstruction-based V2V training and reference-guided synthesis, followed by post-filtering and comparative selection.}
    \label{fig:data_pipeline}
\end{figure}

\subsection{Data Preparation}

Figure~\ref{fig:data_pipeline} (left) summarizes the data preparation pipeline for \oursa.
Building on the data processing framework of \previous, \oursa retains its five-stage pipeline, \textit{Clipping}, \textit{Filtering}, \textit{Speech Processing}, \textit{Captioning}, and \textit{Embedding}, while refining the clipping methods and filtering taxonomy.
On the one hand, we expand our data sources to enrich the facial expressions and body movements generated by the model. On the other hand, we introduce additional processing and filtering operators to improve the quality of the training data. For video captioning, we find that temporally ordered dense captions are better suited for streaming video generation than the previous structured descriptions. We therefore redesign the captioning method, and incorporat expert models to improve coverage and reduce hallucinations.

\paragraph{Data Collection.} In addition to continuing to expand the livestream/talking-head videos and film/television content used in \previous, \oursa places particular emphasis on collecting high-quality solo dance videos and 2D/3D animation data. These additions help the model learn a broader range of body movements and improve its performance in animated character generation.

\paragraph{Video Clipping.}
\oursa retains the single-shot clipping approach of \previous while improving the detection of subtle edit points. We observe that some videos, particularly vlogs, contain numerous edits that are difficult to detect. For example, unboxing videos often omit the intermediate steps of opening a package, retaining only the beginning and end of the action. However, simply increasing shot detection sensitivity introduces many false positives. To address this issue, we extract frames around each candidate cut and use a vision-language model for a second-stage assessment. This procedure ensures that the resulting clips contain no cuts while keeping the false rejection rate below $2\%$.

\paragraph{Video Filtering.}
\oursa retain the six-dimensional filtering taxonomy from \previous: \textit{subject detection}, \textit{frame cleanliness}, \textit{visual quality}, \textit{content safety}, \textit{shot stability}, and \textit{interactivity}. We also introduce a \textit{high-clarity video selection operator} to meet the stricter training-data requirements for real-time 720p video generation. In addition, we relax the hard filtering criterion for shot stability. Videos with small or smooth camera movements are retained and subsequently processed by a \textit{background stabilization operator} to produce stable backgrounds.

\paragraph{High-Clarity Video Selection.}
Videos from different sources vary in compression strategy and severity, so their actual clarity can differ substantially even at the same nominal resolution. Even at 1080p or higher, heavily compressed videos may suffer from texture loss and compression artifacts. The resolution alone is therefore insufficient for selecting high-quality data. We therefore develop a multidimensional hybrid selection framework that evaluates different video types separately. Specifically, hard thresholds are first applied to resolution and frame rate, after which the interdependencies among resolution, frame rate, codec, pixel format, bit depth, and bitrate are assessed. Technical quality, texture detail, edge sharpness, and compression artifact severity are further evaluated using expert models. These assessments are aggregated into a weighted quality score, and a threshold is applied to select training data that satisfy the desired clarity criteria.


\paragraph{Background Stabilization.}

Mitigating background drift and preserving scene consistency remain key challenges in infinite-length streaming video generation, motivating the use of training videos with static backgrounds or fixed cameras. 
However, videos featuring rich body motion, such as dance videos, often include  camera movements such as orbiting, dolly motion, zooming, and subject tracking. Excluding these videos would reduce motion diversity, whereas retaining them without stabilization could impair the learning of background consistency. To address this dilemma, we introduce a \textit{background stabilization operator}.  The approach is motivated by the observation that, in most dance videos, camera translation is limited and visual changes arise primarily from camera rotation and focal length adjustments. The resulting parallax is typically small, allowing background motion to be approximated by geometric transformations across frames. Videos with large, complex camera movements are therefore excluded, while those with smooth motion are retained for stabilization. For the retained videos, foreground subjects are detected and masked, after which camera transformations are estimated from the remaining background regions through feature extraction and matching. Finally, we apply geometric correction and appropriate cropping to obtain videos with stable backgrounds, followed by a second filtering pass to ensure that the subject remains in the frame.


\paragraph{Data Captioning.}

\oursa preserves the two caption granularities used in \previous, \textit{full-clip} and \textit{chunk-level} captions, while redesigning both the caption representation and the annotation pipeline to emphasize temporal structure. \previous uses structured natural-language descriptions with separate fields for subjects, environments, and camera movements. Although this format explicitly distinguishes different scene components, it provides limited support for representing their interactions, temporal evolution, and synchronization. In contrast, \ours adopts temporally ordered dense captions. Aside from a small set of tags summarizing the overall scene, captions describe events chronologically, specifying their temporal boundaries, constituent actions, and outcomes. For example, a caption may describe an action performed between 2.5~s and 3.8~s together with its consequences. This representation makes temporal and causal relationships more explicit, facilitating the modeling of dependencies across events. Consistent with this design, chunk-level captions are segmented at event boundaries rather than assigned to fixed-duration, speech-aligned intervals. The resulting segments are typically finer-grained, enabling lower response latency and greater consistency during action transitions and adaptation to new instructions. To support this temporally structured representation, the annotation pipeline follows a hierarchical, multi-agent design. Expert models are introduced to perform object detection and speech recognition, producing a factual grounding layer for subsequent caption generation. Building on this layer, the workflow assigns recognition, description, verification, and filtering to dedicated agents, improving annotation quality and consistency through explicit task decomposition and validation. 

\subsection{Method}

\paragraph{Model Overview.}
\label{eq:avator_overview}
\oursa employs an audio-visual joint Diffusion Transformer for reference-conditioned video--audio generation. Let $\bm{r}$ denote the reference image. We associate each video--audio segment $\bm{x}^{i}$ with conditioning information $\bm{c}^{i}$, such as its caption, forming the conditioning sequence $\bm{c}^{1:\infty}=\{\bm{c}^{1},\bm{c}^{2},\ldots\}$. Given $\bm{r}$ and $\bm{c}^{1:N}$, the model jointly predicts the clean video--audio latents for the first $N$ segments:
\begin{equation}
\hat{\bm{x}}_{0}^{1:N}
=
\bm{f}_{\bm{\theta}}^{\mathrm{avatar}}
\left(
\bm{x}_{t}^{1:N},
t,
\bm{r},
\bm{c}^{1:N}
\right),
\end{equation}
where $\bm{x}_{t}^{1:N}$ denotes the noisy joint video--audio latents at diffusion timestep $t$. The reference image is shared across all segments to maintain appearance and identity.

\paragraph{Bidirectional Image- and Reference-to-Video Training.}
To support both image-to-video (I2V) and reference-to-video (R2V) generation within a single model, we jointly train a bidirectional model on the tasks:
\begin{equation}
\hat{\bm{x}}_{0}^{1:N}
=
\bm{f}_{\bm{\theta}}^{\mathrm{bi}}
\left(
\bm{x}_{t}^{1:N},
t,
\bm{r},
\bm{c}^{1:N}
\right).
\end{equation}
For I2V training, $\bm{r}$ represents the first frame of the target video; for R2V training, it represents the provided reference image. In both tasks, we supervise each temporal segment with its corresponding condition $\bm{c}^{i}$, such as its caption, instead of using a single prompt for the entire sequence as in conventional bidirectional models. This segment-wise conditional supervision substantially improves instruction following while preserving the model's original generation quality.

\paragraph{Hybrid Teacher and Diffusion Forcing.}
Starting from the pretrained bidirectional model, we replace its bidirectional
temporal attention with a block-wise causal attention mask, such that each
video-audio segment can only attend to the reference image, the current segment condition, and its valid historical states. For the
$i$-th segment, the causal denoising process is formulated as
\begin{equation}
\hat{\bm{x}}_{0}^{i}
=
\bm{f}_{\bm{\theta}}^{\mathrm{causal}}
\left(
\bm{x}_{t_i}^{i},
t_i,
\bm{r},
\bm{c}^{i},
\bm{x}_{\tau_i}^{<i},
\tau_i
\right),
\end{equation}
where $t_i$ denotes the diffusion timestep for the $i$-th segment, and
$\bm{x}_{\tau_i}^{<i}$ denotes the historical video-audio states at noise
level $\tau_i$. We adopt a hybrid training strategy that combines
Teacher Forcing and Diffusion Forcing~\cite{DF}. Under Teacher Forcing, the model is
conditioned on clean ground-truth historical states. Under Diffusion Forcing, noise is injected into the historical
states at sampled noise levels, training the model to generate under the noisy condition.
The two modes are sampled during training with a predefined
probability. This causal adaptation equips the model with an initial
capability for streaming video-audio generation while improving its
robustness to accumulated errors in the generated history.

\paragraph{Self-Replay Forcing.}
\label{eq:srf}
We introduce Self-Replay Forcing (SRF), an on-policy DMD~\cite{yin2024one,yin2024improved,yin2025slow,zeng2026lpm} that aligns the student's autoregressive rollout distribution with the teacher distribution while preserving cross-block gradient flow.
The model first performs a long autoregressive rollout following
Self-Forcing~\cite{huang2026self}, with the generated blocks and KV caches
detached to avoid retaining the full rollout computation graph.
Here, ``on-policy'' refers to the autoregressive trajectory and historical
contexts generated by the current student model under its inference
procedure.
After the rollout is completed, we sample self-generated trajectory and re-noise its blocks following Diffusion
Forcing.
The noisy video is then processed by a gradient-enabled causal replay,
while the external history inherited from the detached rollout remains
fixed.
DMD supervision is applied to all blocks within the replayed segment.
Because the replay-segment representations remain connected within the same
computation graph, gradients can propagate across block boundaries during
the replay pass without backpropagating through the original rollout.
Following the perceptual regularization used in \previous~\cite{vidus1}, we
additionally apply a perceptual loss to the replayed student outputs to
mitigate mode collapse and preserve generation diversity.
Following \previous, we retain sink blocks, a sliding-window
context, and noisy KV caches throughout training.

Let $\hat{\bm{x}}_0^{1:N}$ denote a detached autoregressive rollout
of the current student, and let $\bm{x}_{t}^{1:N}$ denote its
re-noised counterpart, where
$\bm{t}=(t_{1},\ldots,t_{N})$ specifies the diffusion timestep
for each segment. Under noisy-history conditioning, we optimize
the replayed student outputs using
\begin{equation}
\begin{aligned}
\mathcal{L}_{\mathrm{SRF}}
=
\mathcal{L}_{\mathrm{DMD}}\left(
\bm{f}_{\bm{\theta}}^{\mathrm{causal}}\left(
\bm{x}_{\bm{t}}^{1:N},
t,
\bm{r},
\bm{c}^{1:N}
\right)
\right)
+
\mathcal{L}_{\mathrm{perc}}(\bm{f}_{\bm{\theta}}^{\mathrm{causal}}\left(
\bm{x}_{\bm{t}}^{1:N},
t,
\bm{r},
\bm{c}^{1:N}
\right)).
\end{aligned}
\end{equation}

\paragraph{Preference Optimization}
We apply preference- and reward-based optimization at both the bidirectional and streaming stages to mitigate visual quality degradation and improve instruction following.
At the bidirectional stage, we use diffusion-based Direct Preference Optimization (DPO)~\cite{diffusion_dpo} to enhance visual fidelity, facial expressiveness, motion naturalness, and audio--visual synchronization, thereby providing a stronger teacher for subsequent causal adaptation.
At the streaming stage, we apply Streaming Negative-aware Fine-Tuning (Streaming NFT)~\cite{zheng2026diffusionnft} to the causal backbone using self-generated trajectories constructed following the principle of Self-Replay Forcing.
Specifically, the model first performs a detached autoregressive rollout and then replays sampled trajectory for reward-based optimization.
This procedure aligns the optimization states with the inference-time autoregressive distribution, improving visual quality, motion controllability, and instruction adherence under real-time streaming inference.

\paragraph{Super-Resolution Refiner.}
To recover fine-grained spatial details from the low-resolution latent outputs of the causal backbone, we introduce a one-step super-resolution refiner that operates directly in the latent space.
Inspired by the stage-aware cache design of \textbf{TwinCache} introduced in \previous~\cite{vidus1}, we use asymmetric noise levels for the historical caches of the causal backbone and the Refiner.
Specifically, the backbone attends to a high-noise cache $\hat{\bm{x}}_{\tau_{\mathrm{B}}}^{<i}$ to propagate coarse motion and long-range temporal structure, whereas the Refiner attends to a low-noise high-resolution cache $\hat{\bm{x}}_{\tau_{\mathrm{R}},\mathrm{HR}}^{<i}$, where $\tau_{\mathrm{B}}>\tau_{\mathrm{R}}$.
Given the low-resolution latent $\hat{\bm{x}}_{0,\mathrm{LR}}^{i}$ generated for the $i$-th segment, the Refiner first performs latent-space upsampling and adds noise at a fixed refinement timestep $t_r$, and then predicts the corresponding high-resolution latent:
\begin{equation}
\bm{z}_{t_r}^{i}
=
(1-t_r)
\mathcal{U}\!\left(
\hat{\bm{x}}_{0,\mathrm{LR}}^{i}
\right)
+
t_r\bm{\epsilon}^{i},
\qquad
\bm{\epsilon}^{i}\sim\mathcal{N}(\bm{0},\bm{I}),
\end{equation}
where $\mathcal{U}$ denotes latent-space spatial upsampling, $\bm{\epsilon}^{i}\sim\mathcal{N}(\bm{0},\bm{I})$.
The cache $\hat{\bm{x}}_{\tau_{\mathrm{R}},\mathrm{HR}}^{<i}$ denotes the historical high-resolution context maintained at the low noise level $\tau_{\mathrm{R}}$.
All chunks in a replayed trajectory share the same cache noise level $\tau_{\mathrm{R}}$, while the refinement timestep $t_r$ is fixed across segments.
This asymmetric cache scheduling separates temporal propagation from spatial detail restoration.
The high-noise backbone cache provides a coarse temporal prior that is less sensitive to high-frequency artifacts, thereby stabilizing long-range motion propagation.
In contrast, the low-noise high-resolution cache preserves local appearance and identity cues that are important for spatial refinement.
Consequently, the backbone preserves long-range motion consistency, while the Refiner recovers fine-grained details without disrupting the temporal trajectory.

\subsection{Inference Infrastructure}
\label{sec:avatar-infer}

To deliver fast and efficient inference for \oursa, we adopt and extend selected techniques from TurboDiffusion~\citep{zhang2025turbodiffusion} and TurboServe~\citep{jiang2026turboserve} in an end-to-end inference acceleration framework. We optimize operator execution and memory access, and arrange module execution along a shared timeline. This schedule allows different modules to share GPU resources at different times. Together, these optimizations preserve generation quality while improving inference efficiency and overall resource utilization, enabling real-time, low-latency inference services. The main components are described below.

\paragraph{Efficient Attention.}
In streaming video generation with diffusion models, attention contributes significantly to execution time and computational cost~\citep{zhangefficient}, making it a key bottleneck for low-latency inference. Different layers respond differently to reduced precision and approximate attention computation. We therefore use a layer-wise hybrid attention strategy. For each layer, we choose a suitable method from SageAttention~\citep{zhang2025sageattention,zhang2024sageattention2,zhang2025sageattention2++,zhang2025sageattention3,zhang2026sagebwd}, SpargeAttention~\citep{zhang2025spargeattention,zhang2026spargeattention2}, and Sparse-Linear Attention (SLA)~\citep{zhang2025sla,zhang2026sla2}. We use more aggressive methods in less-sensitive layers and prioritize accuracy in sensitive layers. This strategy reduces attention latency while maintaining generation quality.

\paragraph{Quantized Linear-Layer Acceleration.}
Running linear layers at BF16 or FP16 precision preserves accuracy, but their high computational cost makes real-time inference difficult and leads to high latency. Per-tensor and per-channel quantization methods are faster, but a few outliers can dominate their quantization ranges. As a result, most values are represented with low precision, which can severely degrade output video quality. We therefore develop an accurate and efficient CUDA implementation of per-block W8A8 GEMM~\cite{zhang2025accurate} for linear layers. Fine-grained scaling limits the effect of outliers, while the optimized kernels reduce memory usage and accelerate linear-layer computation. The resulting operator maintains numerical precision and generation quality while reducing inference latency.

\paragraph{Kernel Fusion and Launch Optimization.}
Streaming video inference repeatedly executes many short operators in a stable pattern. In this setting, kernel launches and synchronization add noticeable overhead, while intermediate tensors create extra global-memory traffic. We address these costs in two ways. First, we fuse adjacent operations into custom Triton/CUDA kernels. Fusion reduces the number of kernel launches and cuts global-memory reads and writes for intermediate results. For example, we fuse RMSNorm with selected elementwise operations. Second, we use CUDA Graphs for stable execution sequences. Each graph is captured once and replayed one or more times per inference step. A replay replaces many host-side kernel launches with a single graph launch. Together, these techniques reduce launch overhead, synchronization costs, and global-memory traffic while improving GPU utilization.

\paragraph{Multi-GPU Parallelism.}
Real-time inference must meet strict latency targets under limited compute and memory budgets. We address this challenge with multi-GPU context parallelism. We use Ulysses-style context parallelism~\citep{jacobs2023deepspeed} to divide the workload across GPUs and distribute activation memory. Context parallelism introduces collective communication between devices. We quantize the exchanged tensors to reduce both transfer volume and communication latency. This allows multi-GPU execution to scale more efficiently.

\subsection{Agentic System}
\label{sec:agentic-system}
\begingroup
\setlength{\intextsep}{6pt plus 1pt minus 1pt}

In \oursa, users provide an instruction through text or speech, upload an
initial image, and optionally add reference images
(Figure~\ref{fig:agentic-interface}). Spoken instructions are transcribed
into text. A vision-language model (VLM) agent uses these inputs to generate
prompts and reviews the generated frames to refine subsequent prompts.

\begin{figure}[H]
    \centering
    \captionsetup{skip=4pt}
    \includegraphics{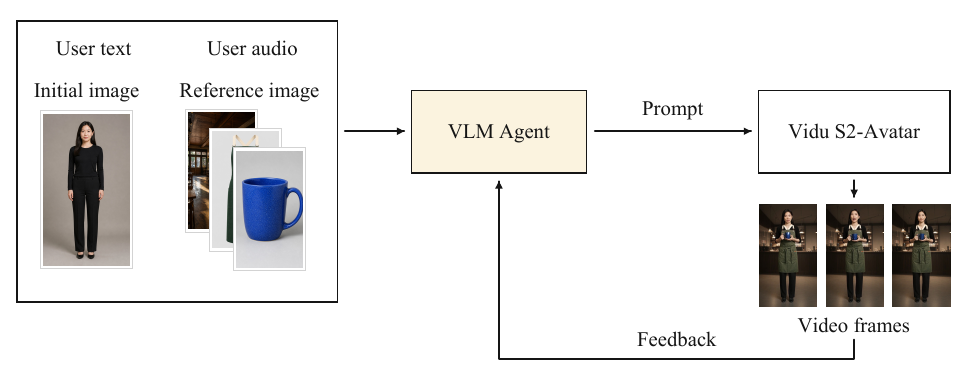}
    \caption{Agentic System Pipeline. The VLM agent reads user instructions and images, writes prompts for \oursa, and checks generated video frames to guide later prompts. The pipeline is schematic.}
    \label{fig:agentic-interface}
\end{figure}

\paragraph{Prompt Generation.}
The prompts separately describe the character's identity and appearance,
its expression, gaze, pose, and actions, and any object that should remain
held. The prompts retain details that the user's instruction does not ask
to change. For example, in the instruction ``pick up a cup, then smile,'' once the cup
has been picked up, the character should continue holding it while smiling.
Subsequent prompts preserve this intended state by explicitly specifying
that the character continues to hold the cup unless the user asks to release
or replace it. The prompt for the smile therefore changes only the expression
description.

\paragraph{Visual Feedback.}
The VLM reviews frames in time order and reports whether the requested
action has finished, has been partly performed, or differs from the request. The system checks confidence
and, when required, agreement across repeated observations. An accepted
completion judgment can lead to a prompt describing the resulting pose or
held object without repeating the action. Incomplete or incorrect actions
can lead to revised instructions; an inconclusive judgment does not confirm
success.

\paragraph{Reference Images.}
The agent first determines whether a reference image shows a handheld
object, a background, or clothing. It then combines the reference content
with the user's request to generate prompts describing how the character
and scene in the initial image should change. This supports object
replacement, background changes, and one-click outfit changes, while
retaining details unrelated to the request. For a scene transition, the
prompt can describe the character leaving the original view and entering
the referenced scene. Figure~\ref{fig:agentic-cases} illustrates object,
scene, and accessory changes.

\paragraph{Taking Off and Putting On Accessories.}
The prompt describes both the movement and whether the accessory should
remain worn afterward. When the user asks to take off a hat and put it back
on, the prompt describes reaching for it, lifting it off, holding it, and
returning it to the head. Later prompts continue to specify the hat as worn
unless the user asks to remove it.

\FloatBarrier
\begin{figure}[H]
    \centering
    \captionsetup{skip=4pt}
    \includegraphics{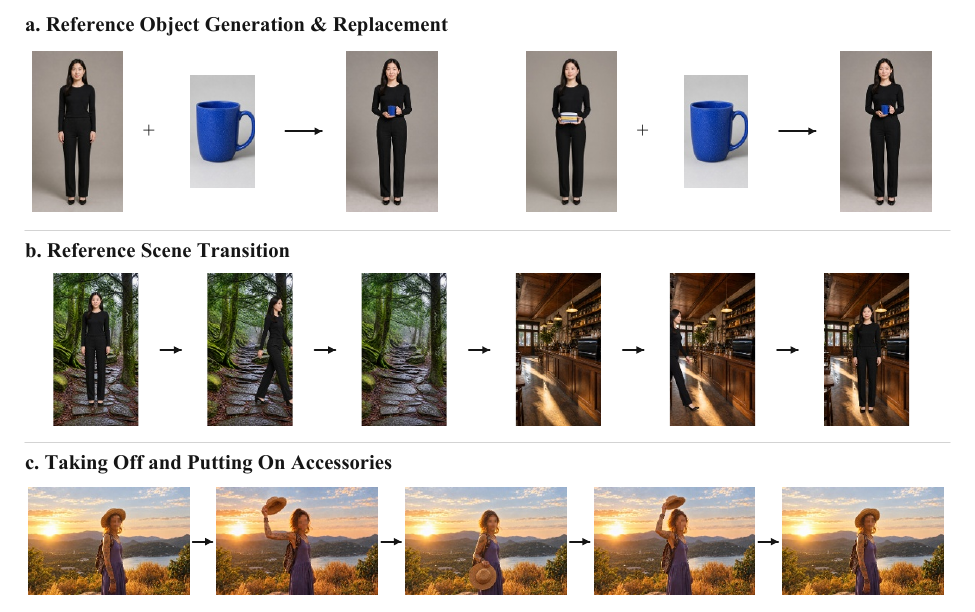}
    \caption{Reference and Accessory Control. (a) Initial images plus a blue-mug reference illustrate object generation (left) and replacement (right). (b) The character leaves the original scene and enters a new one. (c) Taking off and putting back on the same hat. }
    \label{fig:agentic-cases}
\end{figure}

\par
\endgroup

\section{Vidu S2-Editing}

\subsection{Data Preparation}

\paragraph{Data Collection.}
As illustrated in Figure~\ref{fig:data_pipeline} (right), the training data for \ourse is derived from videos filtered through the \oursa data pipeline. We apply stricter filtering criteria and balance the label distribution, yielding a collection of $800,000$ videos with high visual quality and diverse content. From this collection, we sample four mutually disjoint subsets of $200,000$ videos each to construct training data for four editing tasks: \textit{style transfer}, \textit{subject replacement}, \textit{background replacement}, and \textit{virtual try-on}.

\paragraph{Video Editing Data.}

To construct training data for style transfer, we train a conditional video generation model that synthesizes a spatially consistent video from a surface-normal video and a reference image.
Given a raw video clip $v_{\mathrm{raw}}$, we sample one frame from the clip as the reference image $i_{\mathrm{ref}}$ and use a model such as NormalCrafter~\cite{bin2025normalcrafter} to estimate its corresponding surface-normal video $v_{\mathrm{normal}}$.
We then train the conditional model to reconstruct $v_{\mathrm{raw}}$ given $v_{\mathrm{normal}}$ and $i_{\mathrm{ref}}$, thereby learning to preserve the spatial structure and motion specified by the normal condition while following the appearance of the reference image.
To generate style-transfer data, we replace the original reference image with a stylized image while retaining the surface-normal video estimated from the original clip.
The model consequently generates a stylized video that follows the original video's spatial structure and motion.
Our evaluations show that the model generalizes well to more than 50 photorealistic and non-photorealistic styles, including cel shading, cyberpunk, Monet-style painting, sketching, and Chinese gongbi painting, among others.

For other editing tasks, we follow a similar data construction procedure, fine-tuning task-specific video editing models to generate paired training data. We also evaluate several video-editing open-source models, including Bernini~\cite{team2026bernini}, SCAIL-2~\cite{yan2026scail}, Wan2.1 VACE~\cite{jiang2025vace}, SAMA-14B~\cite{zhang2026sama}, and CoinVE-Edit~\cite{long2026coinve}, to supplement the editing data. During data construction, we observe that these models have different strengths across tasks and samples, with no single model consistently producing the best results across all editing scenarios. We therefore apply post-filtering and comparative selection to the candidate outputs from different models, retaining high-quality source--edited video pairs for the final training set.

\paragraph{Data Captioning.}

Each caption identifies the editing task as style transfer, subject replacement, background replacement, or virtual try-on, and describes the attributes of the reference image relevant to that task. For example, style-transfer captions characterize the image’s visual style, while virtual try-on captions describe its clothing attributes. These descriptions define how the reference image should guide the edit. The prompts also explicitly require the model to preserve all source-video content outside the intended edits. To support streaming video editing, captions are generated without feeding the source video to the VLM. The resulting captions thus specify editing objectives and preservation constraints without describing the source video itself, preventing its content from leaking through text conditioning.


\subsection{Method}

\paragraph{Model Overview.}
\ourse employs a Diffusion Transformer (DiT) for instruction-guided video editing conditioned on a source video and an optional reference image. Let $\bm{s}^{1:N}$ denote the source-video condition and $\bm{r}$ the reference-image representation. We represent the remaining conditions as
$\bm{c}^{1:\infty}=\{\bm{c}^{1},\bm{c}^{2},\ldots\}$,
where $\bm{c}^{i}$ denotes the conditioning information for the $i$-th video segment, such as its editing instruction.  For $N$ target segments, the model predicts the clean video latents as
\begin{equation}
\hat{\bm{x}}_{0}^{1:N}
=
\bm{f}_{\bm{\theta}}^{\mathrm{edit}}
\left(
\bm{x}_{t}^{1:N},
t,
\bm{s}^{1:N},
\bm{r},
\bm{c}^{1:N}
\right),
\end{equation}
where $\bm{x}_{t}^{1:N}$ denotes the noisy target-video latents at diffusion timestep $t$. The source video provides spatial and motion context, while the reference image supplies additional appearance information when available. We encode the source video and any reference image as conditioning tokens and concatenate them with the noisy target-video tokens. Condition-specific RoPE encodings represent the spatial and temporal positions of the different token streams.

\paragraph{Bidirectional Video Editing Training.}
We first train a bidirectional model to establish strong visual and motion priors for video editing.
The generated-video tokens interact bidirectionally across the target sequence, while their interaction with the source video is restricted to temporally aligned frame pairs.
Each generated frame exchanges information only with its corresponding source frame, avoiding unrestricted mixing between misaligned source and target contents.
Meanwhile, every generated frame can attend to the reference-image tokens, allowing the reference appearance to be consistently propagated throughout the video.
This frame-aligned attention design preserves the modeling capacity of bidirectional training while making the learned conditioning structure compatible with subsequent causal adaptation.

\paragraph{Causal Streaming Training.}
The causal streaming training procedure for \ourse follows the same
hybrid forcing and Self-Replay Forcing framework described for \oursa.
Starting from the bidirectional editing model, we replace bidirectional
temporal attention with a block-wise causal attention mask and mix Teacher
Forcing with Diffusion Forcing to improve robustness to imperfect historical
contexts.
The model is conditioned on the source video, reference image, editing
prompts, and valid historical states available up to the current segment.
We then apply Self-Replay Forcing by first performing a detached
autoregressive rollout, followed by re-noising and differentiable replay of the self-generated trajectory.
DMD supervision is applied to the replayed blocks, allowing the model to learn
from inference-time states without backpropagating through the original
rollout.
This procedure equips \ourse with an capability for
real-time streaming video editing while improving temporal consistency and
instruction adherence.

\subsection{Inference Infrastructure}

For \ourse, the source video stream is produced continuously on the user side. The model must process each incoming video frame promptly to keep the edited output synchronized with user actions. In real-time editing tasks, the same inference delay is more noticeable than in generation tasks because users directly compare their actions with the edited output. Based on the inference optimizations developed for \oursa, which are described in \reference{sec:avatar-infer}, we further improve how GPU resources are scheduled across modules. The following design reduces idle GPU resources and lowers end-to-end latency.

\textbf{Inter-Module Scheduling.}
Modules are active at different stages of streaming video inference. A static placement policy reserves a fixed GPU group for each module, even when the module is idle. This wastes GPU resources and increases end-to-end latency when GPU capacity is limited. We instead use fine-grained scheduling to coordinate the VAE encoder, backbone, refiner, and VAE decoder on a shared timeline. When the backbone or refiner does not need certain GPUs, the VAE encoder or decoder can reuse them. This time-based sharing reduces dedicated GPU requirements, improves overall resource utilization, and lowers inference latency.

\section{Real-Time Spatial Video Generation and Editing}
\label{sec:stereoscopic-video}

Spatial video presents a slightly different image to each eye.
This lets viewers see depth, scale, and the positions of objects in a scene.
Compared with flat video, it creates a stronger sense of presence. Characters
and scenes can feel as if they are around the user instead of on a screen.
This creates a more immersive entertainment experience.

\subsection{Real-Time Spatial Video Generation}

\paragraph{Motivation.}
We want real-time video generation to offer a more immersive entertainment experience. \oursa can
already generate monocular video in real time, with rich expressions, motion,
and interactive responses. However, flat video gives a weaker sense of space
and presence. We therefore aim to convert its streaming output into
synchronized left and right views while preserving the real-time
responsiveness needed for continuous interaction.

\paragraph{Generation Pipeline.}
Our generation pipeline has two stages. First, \oursa uses the current text, audio, visual references, other controls, and
video history to generate the next monocular video chunk. Second, the conversion stage estimates per-frame depth and maps it to
horizontal disparity~\citep{zhang20113d}. Treating the monocular frame as a center view, it
warps the image in opposite horizontal directions to synthesize the left
and right views.
Warping leaves holes along object boundaries and newly exposed regions~\citep{zhao2024stereocrafter,huang2026dreamstereo}.
For low-latency streaming, we use lightweight processing to mitigate artifacts near depth discontinuities and fill holes in disoccluded regions. We temporally stabilize
depth estimates to reduce fluctuations in perceived depth. Finally, the two views are synchronized as
streaming video chunks.
The streaming spatial video allow users to
perceive the distance between the character and the background, the scale of
objects, and the depth of the scene, making the generated character feel more
present in the user's space. This streaming design preserves the capability of \oursa to support real-time character interaction in spatial video form.

\subsection{Real-Time Spatial Video Editing}

\paragraph{Motivation.}
Many existing videos are available only in monocular form. Spatial video is
costly to produce and difficult to modify once captured or generated.
These limitations motivate two workflows: editing ordinary
monocular video and then converting it into synchronized left- and right-eye views, or directly editing existing spatial video.

\paragraph{Editing Pipeline.}
Our real-time spatial video editor is based on \ourse and supports
two workflows. (1) \textit{Monocular input.} We first edit the incoming
monocular stream with \ourse and then pass the edited output through
the same spatial video
conversion pipeline used for \oursa, producing synchronized left- and
right-eye views. (2) \textit{Stereoscopic input.} For spatial video, we join
the left and right views horizontally to form one wide video stream. \ourse edits both views in one pass with the same visual references, then
we split the output back into left and right views.

Both workflows can send synchronized output to the left- and right-eye
displays. Whether the input is monocular or stereoscopic, users can perform
style transfer, virtual try-on, character replacement, and background
replacement. These edits can be applied either to video content or to the
physical world as seen through a live camera or headset passthrough.

\subsection{Discussion}
The generation and editing pipelines above demonstrate that \ours can extend real-time monocular video to spatial-video experiences. However, for practical deployment, this task requires higher resolution and lower latency than conventional real-time monocular video generation. When viewed through a headset, spatial video often occupies a large portion of the user's field of view and therefore requires high resolution to maintain visual comfort. Low latency is equally important. For example, in camera-based video passthrough mode, high latency can make head motion and the displayed view fall out of sync. Achieving low end-to-end latency while maintaining high visual quality remains a key technical challenge.

Beyond these challenges, a promising future direction is to extend the current system from
fixed-view spatial video to panoramic spatial video, allowing users to turn their heads and freely
explore generated scenes. We believe this capability could also support a metaverse whose visual content is generated in real time. Scenes, characters, and events could change with user interaction. Such real-time video generation could substantially reduce the need for traditional 3D modeling, material creation, animation, and scene building.

\section{Experiments}
\label{sec:experiments}

We evaluate \ours on two complementary tasks: streaming digital character generation and streaming video editing. \reference{sec:exp-setup} introduces the task definitions, benchmark suite, comparison systems, and evaluation protocols. \reference{sec:exp-public} reports results against open-source systems on public benchmarks, with separate analyses for digital character generation and video editing. \reference{sec:exp-gsb} assesses commercial systems on internal benchmarks using randomized Good/Same/Bad (GSB) pairwise comparisons. Finally, \reference{sec:exp-qualitative} examines representative cases of identity preservation, temporal consistency, robustness during long-horizon streaming and promising spatial video generation ability.

\subsection{Experimental Setup}
\label{sec:exp-setup}

This section defines the task and benchmark suite in \reference{sec:exp-tasks}, summarizes the comparison systems and access conditions in \reference{sec:exp-baselines}, and specifies the public-benchmark and GSB protocols in \reference{sec:exp-protocols}.

\subsubsection{Tasks and Benchmark}
\label{sec:exp-tasks}

We study two tasks. For digital character generation, \oursa generates an audio-driven streaming character from an audio track, a reference identity image, and a textual condition. For video editing, \ourse performs instruction-guided and reference-conditioned edits in the streaming setting.

For digital character generation, we use StreamAV-Bench~\citep{liu2026streamavbench} for public evaluation and an internal benchmark for commercial-system comparison. StreamAV-Bench contains 160 scenarios in each of its Progressive and Interactive tracks. The Interactive Track receives one runtime update every 30 seconds for five updates and runs for up to 180 seconds. Our internal digital character benchmark covers streaming quality, long-horizon stability, runtime updates, interruption and recovery, audio--video synchronization, system responsiveness, and cost. It follows the StreamAV-Bench evaluation protocol wherever the measurements are shared.

For video editing, we use OpenVE-Bench~\citep{he2025openve3m}, Sparkle-Bench~\citep{zeng2026sparkle}, RefVIE-Bench~\citep{lin2026kiwiedit}, and the ViViD test set~\citep{fang2024vivid}. OpenVE-Bench and Sparkle-Bench evaluate instruction-guided editing; RefVIE-Bench adds a reference image for subject and background replacement; and the ViViD test set evaluates unpaired virtual try-on. Our internal editing benchmark covers style transfer, virtual try-on, subject replacement, and background replacement across diverse subjects, references, motions, and scene layouts. We also construct a long-horizon set with clips at the 10-minute level to test identity preservation, edited-region stability under motion and occlusion, and temporal stability of generated backgrounds.

\subsubsection{Baselines}
\label{sec:exp-baselines}

For the public digital character evaluation, we compare \oursa with the systems listed in Table~\ref{tab:streamav}: PixVerse R1~\citep{pixverse2026r1}, HappyOyster~\citep{alibaba2026happyoyster}, OmniForcing~\citep{su2026omniforcing}, Odyssey-2~\citep{odyssey2025odyssey2}, Self-Forcing~\citep{huang2026self}, LongLive~\citep{yang2025longlive}, SWIFT~\citep{tan2026swift}, IAMFlow~\citep{liu2026iamflow}, SoulX-FlashTalk~\citep{shen2025soulxflashtalk}, AptAvatar~\citep{zhang2026aptavatar}, Live Avatar~\citep{huang2025live}, AvatarForcing~\citep{cui2026avatarforcing}, and Hallo-Live~\citep{li2026hallolive}. The internal duration-stratified curves in Figure~\ref{fig:avatar-duration-curves} compare \oursa with Vidu S1~\citep{vidus1}, Runway, HeyGen~\citep{heygen2026avatar}, and PixVerse~\citep{pixverse2026r1}. For public video editing, we compare \ourse with the systems appearing in Tables~\ref{tab:sparkle}--\ref{tab:vivid}: Bernini-R 1.3B and Bernini-R 14B~\citep{team2026bernini}, Kiwi-Edit 5B~\citep{lin2026kiwiedit}, VInO~\citep{chen2026vino}, OmniWeaving~\citep{pan2026omniweaving}, LiveEdit~\citep{wang2026liveedit}, VACE 1.3B~\citep{jiang2025vace}, JoyAI-Video-Edit~\citep{xiao2026joyaivideoedit}, StreamDiffusionV2~\citep{feng2025streamdiffusionv2}, XMax-X2.0~\citep{xmax_x20}, Decart-Lucy2.5~\cite{decart_lucy25}, StableVITO~\citep{kim2023stableviton}, OOTDiffusion~\citep{xu2024ootdiffusion}, IDM-VTON~\citep{choi2024idmvton}, StableVITON~\citep{kim2023stableviton}, StableVITON+AM, OOTDiffusion+AM, IDM-VTON+AM, ViViD~\citep{fang2024vivid}, and CatV$^2$TON~\citep{zheng2024catvton}. The internal editing GSB comparison in Figure~\ref{fig:editing-gsb} includes XMax-X2.0 and Decart-Lucy2.5. Face-only APIs are evaluated in a separate lip-sync and latency track and are not merged with full-body systems.

\subsubsection{Metrics}
\label{sec:exp-protocols}

For public benchmarks, we use identical inputs, preprocessing, temporal sampling, evaluators, and scoring rubrics. Every system receives the same manifest, audio, and reference image. We reuse a published result only when the source reports the same benchmark, split, and metric; otherwise, we omit the entry. For metrics inherited from a benchmark, we retain the benchmark's original directionality and procedure.

For digital character generation, we report the metrics defined by StreamAV-Bench~\citep{liu2026streamavbench}. Visual Aesthetics (VA) measures perceptual visual appeal with the LAION Aesthetic Predictor~\citep{laion2022aesthetic}. Visual Quality (VQ) is a Gemini-based assessment of visual fidelity, subject integrity, motion naturalness, and visible artifacts. Production Quality (PQ) measures the perceptual quality and production fidelity of the generated audio with AudioBox Aesthetics~\citep{tjandra2025audiobox}, whereas Audio Quality (AQ) uses the Gemini judge to assess audio naturalness and audible artifacts. Audio--Visual Alignment (AVAlign) measures semantic correspondence between the generated audio and video using ImageBind similarity~\citep{girdhar2023imagebind}; Audio--Visual Synchronization (AVSync) estimates the temporal synchronization error with Synchformer~\citep{iashin2024synchformer}, so lower values are better. Audio Instruction Fulfillment (AIF) measures how completely the audio-side requirements in the case-specific checklist are realized. Subject Consistency (SC) and Background Consistency (BC) follow the long-horizon protocol of StreamAV-Bench and quantify the temporal stability of the subject appearance and background scene, respectively. The Gemini-based metrics use the benchmark's ordinal rating rubric, whereas VA, PQ, AVAlign, AVSync, SC, and BC retain their native metric scales. FVD and FID are computed with the benchmark tools following their original definitions~\citep{unterthiner2018towards,heusel2017gans}; they measure distributional differences between generated and reference video or image features, with lower values indicating closer distributions.

For video editing, the public benchmarks retain their task-specific definitions. OpenVE-Bench~\citep{he2025openve3m} reports Global Style (GS), which evaluates whether the requested global appearance or style is achieved, Background Change (BC), which evaluates the correctness of the background edit, and an Overall score that aggregates the benchmark's judged editing quality. Sparkle-Bench~\citep{zeng2026sparkle} reports global instruction compliance (Ins.), global visual quality (Vis.), foreground instruction adherence (FgIn.), foreground motion preservation (FgMo.), background dynamics (BgDy.), and background visual quality (BgVi.). RefVIE~\citep{lin2026kiwiedit} evaluates reference-conditioned editing through reference fidelity, matting or boundary quality, visual harmony, and temporal consistency; its reported RefVIE Overall is the benchmark-defined aggregate of these dimensions. Joint Overall is the aggregate specified by the joint OpenVE--RefVIE protocol. For the unpaired virtual try-on task, VFID$_I$ uses I3D and 3D-ResNetXt101 video features to measure the distance between generated and reference video distributions, thereby reflecting both visual quality and temporal coherence; lower values are better.

For internal benchmarks, we use inputs and randomize the presentation order of every pair. Twenty professionally trained evaluators with relevant backgrounds in computer vision, computer graphics, video production, or visual quality assessment perform the pairwise comparisons. Before formal annotation, they complete calibration rounds with annotated examples and a written rubric; the calibration is used to align the interpretation of motion quality, visual quality, consistency, audio--video synchronization, and semantic compliance. Evaluators inspect the same source inputs and synchronized outputs and select Good when one system is clearly preferred, Same when the difference is not perceptually meaningful, and Bad when the other system is preferred. GSB denotes these Good, Same, and Bad outcomes. Any aggregate win rate specifies its denominator and tie treatment. We retain the input pair, random seed, model version, service region, and annotation record for every sample. Product-page claims about frame rate or latency do not replace measured results.

\begin{table*}[!t]
\centering
\normalsize
\caption{Results on StreamAV-Bench. ``--'' denotes an unavailable measurement. \textbf{Bold} indicates the best value in each column, \underline{underlining} indicates the second-best value, and \tableshading{cvprblue!20}{dark-blue shading} marks our method. $\uparrow$/$\downarrow$ denote higher/lower is better.}
\label{tab:streamav}
\begin{tabularx}{\textwidth}{l*{9}{>{\centering\arraybackslash}X}}
\toprule
\rowcolor{cvprblue!12}
Model & VA$\uparrow$ & VQ$\uparrow$ & PQ$\uparrow$ & AQ$\uparrow$ & AVAlign$\uparrow$ & AVSync$\downarrow$ & AIF$\uparrow$ & SC$\uparrow$ & BC$\uparrow$ \\
\midrule
SoulX-FlashTalk & .281 & 2.515 & 7.091 & 3.006 & .121 & \underline{.648} & 2.499 & .946 & .964 \\
AptAvatar & .458 & 2.040 & 7.075 & 2.582 & .099 & 1.284 & 2.448 & .860 & .923 \\
OmniForcing & .554 & 2.374 & 6.351 & 2.528 & .119 & 1.423 & 2.343 & .962 & .953 \\
Hallo-Live & .574 & -- & 5.396 & -- & .102 & 1.353 & -- & .991 & .977 \\
Self-Forcing & .585 & 2.753 & 6.453 & 2.623 & .256 & .919 & 2.810 & .981 & .969 \\
IAMFlow & .602 & 2.840 & 6.415 & 2.625 & .268 & 1.016 & 2.789 & .986 & .973 \\
LongLive & .605 & 2.768 & 6.459 & 2.696 & \underline{.272} & .969 & 2.814 & .986 & .973 \\
SWIFT & .605 & 2.735 & 6.483 & 2.680 & .270 & .970 & \underline{2.887} & .986 & .973 \\
AvatarForcing & .639 & -- & 7.070 & -- & .089 & 1.041 & -- & .987 & .976 \\
PixVerse R1 & .529 & 2.428 & 6.348 & 2.810 & .234 & .855 & 2.511 & .907 & .930 \\
Odyssey-2 & .531 & 2.820 & 6.426 & 2.659 & .260 & .946 & 2.778 & .976 & .966 \\
HappyOyster & .529 & 2.785 & 6.817 & \underline{3.157} & .206 & 1.044 & 2.692 & .904 & .934 \\
Live Avatar & \underline{.661} & \underline{3.295} & \underline{7.133} & 3.079 & .116 & 1.145 & 2.745 & \underline{.997} & \underline{.989} \\
\rowcolor{cvprblue!20}
\textbf{Vidu S2-Avatar} & \textbf{.687} & \textbf{3.370} & \textbf{7.138} & \textbf{3.286} & \textbf{.353} & \textbf{.617} & \textbf{2.985} & \textbf{.998} & \textbf{.993} \\
\bottomrule
\end{tabularx}
\end{table*}

\subsection{Public-Benchmark Results}
\label{sec:exp-public}

This section compares \ours with open-source systems under the unified protocols in \reference{sec:exp-protocols}. We first report digital character generation results in \reference{sec:exp-public-avatar}, followed by video editing results in \reference{sec:exp-public-editing}.

\subsubsection{Digital-Character Generation}
\label{sec:exp-public-avatar}

\paragraph{StreamAV-Bench~\citep{liu2026streamavbench}.}
Table~\ref{tab:streamav} reports the available Gemini-MLLM subset. \oursa achieves the best value in every reported metric, and the gains are distributed across the complementary dimensions defined by StreamAV-Bench rather than concentrated in a single aspect. The visual metrics (VA and VQ) indicate stronger perceptual appeal and visual fidelity, while the audio metrics (PQ and AQ) show that the generated speech remains clean and natural. The improvement is especially important for an audio-driven avatar: AVAlign and AVSync jointly indicate tighter semantic coupling and more reliable temporal synchronization between speech and facial motion, and AIF further reflects stronger fulfillment of the audio-side instructions. Finally, the near-ceiling SC and BC scores suggest that the identity and scene structure remain stable over long rollouts. Taken together, these results point to a more balanced streaming system that preserves visual quality while maintaining audio quality, cross-modal coordination, and long-horizon continuity.

\begin{table*}[!t]
\centering
\normalsize
\caption{Sparkle-Bench results. Ins., Vis., FgIn., FgMo., BgDy., and BgVi. denote global instruction, global visual quality, foreground instruction, foreground motion, background dynamics, and background visual quality. \textbf{Bold} indicates the best value in each column, \underline{underlining} indicates the second-best value, \tableshading{black!5}{light-gray shading} denotes offline models, \tableshading{cvprblue!8}{pale-blue shading} denotes streaming models, and \tableshading{cvprblue!20}{darker-blue shading} highlights our streaming method. $\uparrow$/$\downarrow$ denote higher/lower is better.}
\label{tab:sparkle}
\begin{tabularx}{\textwidth}{l*{7}{>{\centering\arraybackslash}X}}
\toprule
\rowcolor{cvprblue!12}
Model & Overall$\uparrow$ & Ins.$\uparrow$ & Vis.$\uparrow$ & FgIn.$\uparrow$ & FgMo.$\uparrow$ & BgDy.$\uparrow$ & BgVi.$\uparrow$ \\
\midrule
\rowcolor{black!5}
Bernini-R 1.3B & 3.43 & 3.64 & 3.18 & 3.48 & 3.53 & 3.24 & 3.51 \\
\rowcolor{black!5}
Bernini-R 14B & 3.50 & 3.70 & 3.22 & 3.60 & 3.69 & 3.24 & 3.59 \\
\rowcolor{black!5}
Kiwi-Edit 5B & 3.57 & 3.84 & \underline{3.33} & 3.76 & 3.81 & 3.00 & 3.68 \\
\rowcolor{black!5}
VInO & 3.39 & 3.62 & 3.20 & 3.43 & 3.56 & 3.08 & 3.47 \\
\rowcolor{black!5}
OmniWeaving & 3.39 & 3.66 & 3.14 & 3.51 & 3.65 & 2.92 & 3.47 \\
\rowcolor{black!5}
VACE 1.3B & 2.18 & 2.22 & 2.17 & 2.20 & 2.21 & 2.11 & 2.19 \\
\midrule
\rowcolor{cvprblue!8}
LiveEdit & 3.31 & 3.60 & 3.03 & 3.43 & 3.58 & 3.01 & 3.22 \\
\rowcolor{cvprblue!8}
JoyAI-Video-Edit & 3.48 & 3.74 & 3.22 & 3.66 & 3.74 & 3.09 & 3.45 \\
\rowcolor{cvprblue!8}
StreamDiffusionV2 & 2.01 & 2.01 & 2.01 & 2.00 & 2.00 & 2.01 & 2.01 \\
\rowcolor{cvprblue!8}
XMax-X2.0 & 3.03 & 3.24 & 2.82 & 2.96 & 3.23 & 2.91 & 3.01 \\
\rowcolor{cvprblue!8}
Decart-Lucy2.5 & \underline{3.67} & \underline{3.91} & 3.31 & \underline{3.86} & \underline{3.91} & \underline{3.30} & \underline{3.74} \\
\rowcolor{cvprblue!20}
\textbf{\ourse} & \textbf{3.74} & \textbf{4.00} & \textbf{3.34} & \textbf{3.98} & \textbf{4.00} & \textbf{3.37} & \textbf{3.76} \\
\bottomrule
\end{tabularx}
\end{table*}

\begin{table}[!t]
\centering
\caption{Joint OpenVE and RefVIE evaluation. ``--'' denotes an unavailable measurement. \textbf{Bold} indicates the best value in each column, \underline{underlining} indicates the second-best value, \tableshading{black!5}{light-gray shading} denotes offline models, \tableshading{cvprblue!8}{pale-blue shading} denotes streaming models, and \tableshading{cvprblue!20}{darker-blue shading} highlights our streaming method. $\uparrow$/$\downarrow$ denote higher/lower is better.}
\label{tab:openve-refvie}
\normalsize
\begin{tabularx}{\textwidth}{l*{5}{>{\centering\arraybackslash}X}}
\toprule
\rowcolor{cvprblue!12}
Model & \makecell{OpenVE\\GS$\uparrow$} & \makecell{OpenVE\\BC$\uparrow$} & \makecell{OpenVE\\Ovr.$\uparrow$} & \makecell{RefVIE\\Ovr.$\uparrow$} & \makecell{Joint\\Ovr.$\uparrow$} \\
\midrule
\rowcolor{black!5}
Bernini-R 1.3B & 3.24 & 3.68 & 3.46 & 2.89 & 3.32 \\
\rowcolor{black!5}
Bernini-R 14B & \underline{4.30} & \underline{4.10} & \underline{4.20} & 3.09 & \underline{3.92} \\
\rowcolor{black!5}
Kiwi-Edit 5B & 3.67 & 2.82 & 3.24 & 2.93 & 3.16 \\
\rowcolor{black!5}
OmniWeaving & 4.28 & 2.83 & 3.55 & 3.08 & 3.43 \\
\rowcolor{black!5}
VInO & 4.24 & 1.83 & 3.02 & 2.59 & 2.91 \\
\rowcolor{black!5}
VACE 1.3B & 2.15 & 1.10 & 1.62 & 2.01 & 1.72 \\
\midrule
\rowcolor{cvprblue!8}
XMax-X2.0 & 3.17 & 1.77 & 2.46 & 3.16 & 2.64 \\
\rowcolor{cvprblue!8}
Decart-Lucy2.5 & -- & -- & -- & \underline{3.71} & -- \\
\rowcolor{cvprblue!20}
\textbf{\ourse} & \textbf{4.71} & \textbf{4.14} & \textbf{4.42} & \textbf{3.78} & \textbf{4.26} \\
\bottomrule
\end{tabularx}
\end{table}

\subsubsection{Video Editing}
\label{sec:exp-public-editing}

\paragraph{Sparkle-Bench~\citep{zeng2026sparkle}.}
\ourse achieves the highest Overall score (3.74) and the highest scores for global instruction (4.00), global visual quality (3.34), foreground instruction (3.98), foreground motion (4.00), background dynamics (3.37), and background visual quality (3.76) in Table~\ref{tab:sparkle}. These results show that \ourse performs competitively across both foreground- and background-oriented criteria.



\paragraph{OpenVE and RefVIE~\citep{he2025openve3m,lin2026kiwiedit}.}
As shown in Table~\ref{tab:openve-refvie}, \ourse achieves the highest
scores across all reported metrics, with a Joint Overall score
of 4.26, surpassing the strongest offline baseline, Bernini-R 14B,
by 0.34. On OpenVE, our model leads in both Global Style (4.71)
and Background Change (4.14), achieving an Overall score of 4.42.
On RefVIE, it attains an Overall score of 3.78, outperforming
the streaming baseline Decart-Lucy2.5 by 0.07. These results
demonstrate that \ourse supports streaming video editing while
outperforming the evaluated offline and streaming baselines
on these benchmarks.


\paragraph{ViViD Test Set~\citep{fang2024vivid}.}
As shown in Table~\ref{tab:vivid}, \ourse achieves a VFID$_I$ of 9.9515 on the unpaired virtual try-on benchmark, outperforming CatV$^2$TON (19.5131) and ViViD (21.8032) and indicating closer alignment between the generated and reference video distributions.

\begin{table}[!t]
\centering
\caption{Unpaired virtual try-on evaluation on the ViViD test set. \textbf{Bold} indicates the best value, \underline{underlining} indicates the second-best value, \tableshading{cvprblue!20}{light-blue shading} marks our method. $\uparrow$/$\downarrow$ denote higher/lower is better.}
\label{tab:vivid}
\normalsize
\begin{tabularx}{\textwidth}{>{\raggedright\arraybackslash}X>{\centering\arraybackslash}X}
\toprule
\rowcolor{cvprblue!12}
Method & VFID$_I\downarrow$ \\
\midrule
StableVITO & 36.8985 \\
OOTDiffusion & 35.3170 \\
IDM-VTON & 25.4972 \\
StableVITON+AM & 22.0262 \\
OOTDiffusion+AM & 23.3938 \\
IDM-VTON+AM & 22.5881 \\
ViViD & 21.8032 \\
CatV$^2$TON & \underline{19.5131} \\
\midrule
\rowcolor{cvprblue!20}
\textbf{\ourse} & \textbf{9.9515} \\
\bottomrule
\end{tabularx}
\end{table}

\begin{figure*}[!t]
\centering
\includegraphics[width=0.8\textwidth]{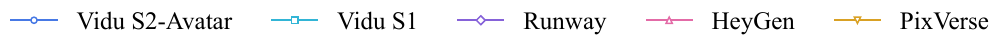}\par
\vspace{2pt}
\begingroup
\setlength{\tabcolsep}{0pt}
\begin{tabular}{@{}>{\centering\arraybackslash}p{0.19\textwidth}@{\hspace{0.0125\textwidth}}>{\centering\arraybackslash}p{0.19\textwidth}@{\hspace{0.0125\textwidth}}>{\centering\arraybackslash}p{0.19\textwidth}@{\hspace{0.0125\textwidth}}>{\centering\arraybackslash}p{0.19\textwidth}@{\hspace{0.0125\textwidth}}>{\centering\arraybackslash}p{0.19\textwidth}@{}}
\includegraphics[width=0.19\textwidth]{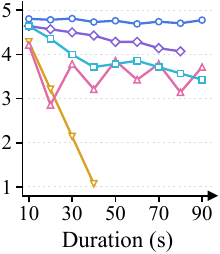} &
\includegraphics[width=0.19\textwidth]{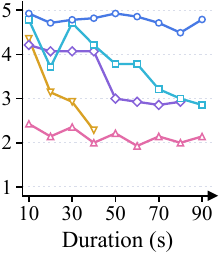} &
\includegraphics[width=0.19\textwidth]{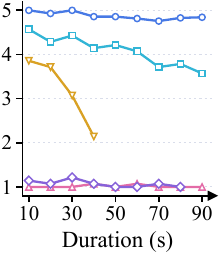} &
\includegraphics[width=0.19\textwidth]{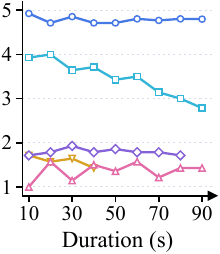} &
\includegraphics[width=0.19\textwidth]{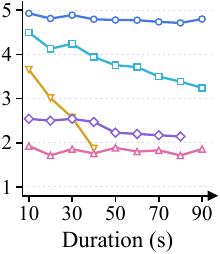} \\[-2pt]
\footnotesize (a) consistency &\footnotesize  (b) video quality & \footnotesize (c) motion quality & {\footnotesize (d) emotional expression} &\footnotesize  (e) overall quality
\end{tabular}
\endgroup
\caption{Duration-stratified mean ratings for digital character generation. Absent tail points indicate unavailable ratings.}
\label{fig:avatar-duration-curves}
\end{figure*}

\subsection{Complementary Human Preference Evaluation}
\label{sec:exp-gsb}

Public benchmarks enable reproducible model comparisons under standardized settings.
However, many existing benchmarks rely heavily on automated or model-based evaluators, which may not fully capture perceptual artifacts that emerge over long video sequences, such as identity drift, temporal inconsistency, and accumulated generation errors.
To complement these evaluations and provide a more comprehensive assessment, we conduct randomized paired human-preference comparisons using the GSB protocol described in \reference{sec:exp-protocols}.
We report the results separately for digital-character generation and video editing.

\paragraph{Digital-Character Generation.}

Commercial digital character systems are evaluated on our internal benchmark with 160, 320, and 480~ms audio chunks, interruption and recovery tests, five-minute stability tests, end-to-end latency, and cost. 

Figure~\ref{fig:avatar-duration-curves} examines how perceived quality evolves as generation extends from 10 to 90 seconds. This duration-stratified view complements clip-level GSB judgments because a streaming model can produce strong short clips yet accumulate identity drift, motion discontinuities, or loss of expressiveness over time. Following the emphasis on long-horizon audio--visual stability in StreamAV-Bench and the separation of visual quality from temporal consistency in VBench~\citep{liu2026streamavbench,huang2023vbench}, we report mean ratings for overall quality, consistency, video quality, motion quality, and emotional expression on a common 1--5 scale.

\oursa consistently achieves the highest ratings across all five dimensions over the evaluated duration range.
Its consistency remains high up to 90 seconds, suggesting that subject identity and temporal structure are preserved during longer streams.
The video- and motion-quality curves remain similarly strong as the duration increases, indicating sustained perceptual fidelity and coherent motion.
Emotional-expression scores also remain high at longer durations, showing that \oursa maintains expressive facial and body motion while preserving long-horizon stability.
Taken together, these curves indicate a broad advantage in long-horizon generation, spanning visual fidelity, motion coherence, identity preservation, and expressive behavior, rather than a gain limited to short clips.
These duration-stratified ratings complement the paired GSB comparisons by characterizing how generation quality evolves with stream duration.


\begin{figure*}[!t]
\centering
\includegraphics[width=0.48\textwidth]{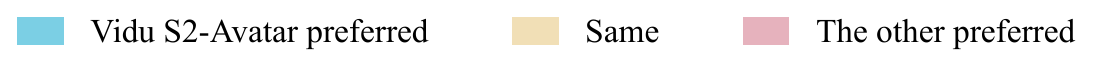}\par\vspace{5pt}
\includegraphics[width=0.98\textwidth]{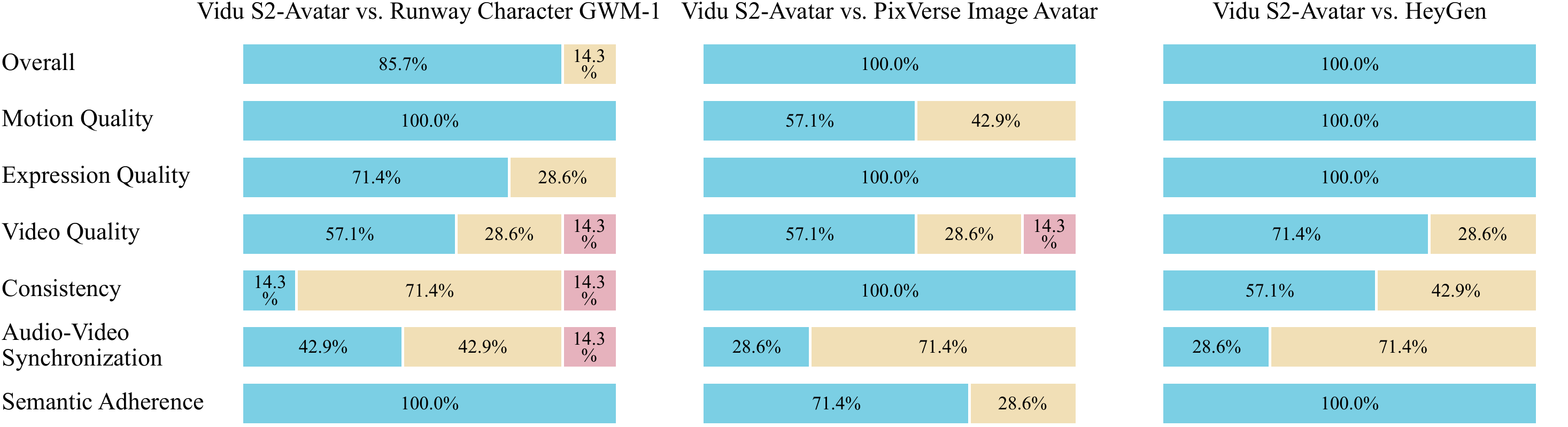}
\caption{GSB human-preference comparison for streaming digital character generation. The figure compares \oursa with Runway Character GWM-1, PixVerse Image Avatar, and HeyGen across overall quality, motion quality, expression quality, video quality, consistency, audio--video synchronization, and semantic adherence. Percentages partition the available paired judgments into \oursa preferred, Same, and the other system preferred; unavailable outcome categories are omitted from the corresponding bar.}
\label{fig:avatar-gsb-closed}
\end{figure*}

\begin{figure*}[!t]
\centering
\includegraphics[width=0.48\textwidth]{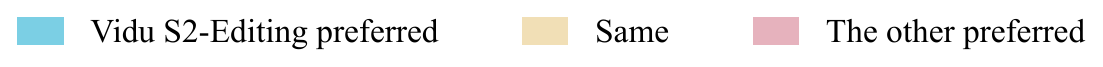}\par\vspace{5pt}
\includegraphics[width=0.98\textwidth]{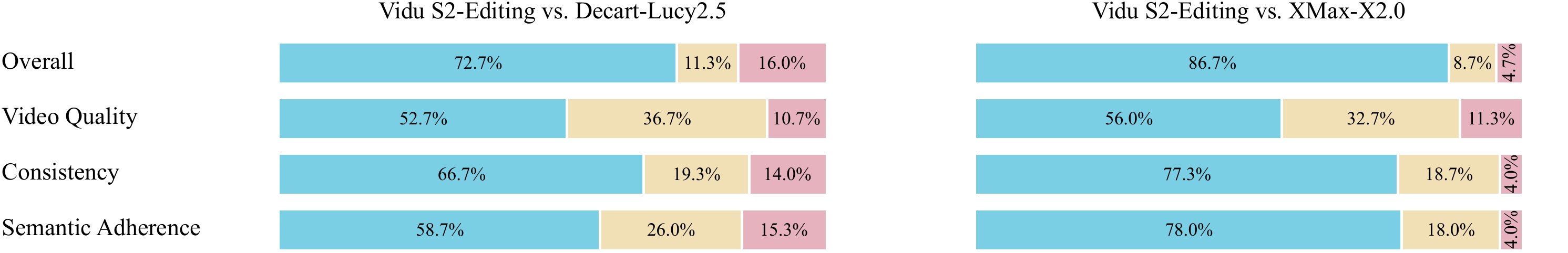}
\caption{GSB human-preference comparison between \ourse and two commercial systems on the internal editing benchmark. Each bar partitions paired judgments into \ourse preferred, Same, and the other system preferred for four criteria.}
\label{fig:editing-gsb}
\end{figure*}

Figure~\ref{fig:avatar-gsb-closed} demonstrates a consistent advantage of \oursa over three closed-source systems across overall quality, motion, expression, semantic adherence, and temporal consistency.
The strongest gains are observed in overall quality, where \oursa is preferred in 85.7\% of the comparisons against Runway Character GWM-1 and in 100\% of the comparisons against both PixVerse Image Avatar and HeyGen.
This advantage is also reflected in motion and expression quality: \oursa receives unanimous preference over Runway and HeyGen for both dimensions, and achieves 100\% preference for expression quality over PixVerse.
For semantic adherence, \oursa achieves preference rates ranging from 71.4\% to 100\%, indicating reliable instruction following across different systems.
The video-quality comparisons further show a consistent preference for \oursa, with preference rates between 57.1\% and 71.4\%.
The temporal-consistency results are particularly favorable, including unanimous preference over PixVerse and a majority preference over HeyGen; in the comparison with Runway, \oursa is preferred or judged perceptually equivalent in 85.7\% of the cases.
For audio--visual synchronization, \oursa is either preferred or judged perceptually equivalent in all comparisons against PixVerse and HeyGen, and is directly preferred in 42.9\% of the comparisons against Runway.
Overall, these results show that \oursa provides broad and consistent improvements in perceptual quality, motion fidelity, expression, semantic adherence, and long-horizon stability while maintaining strong audio--visual synchronization.




\paragraph{Video Editing.}

The internal editing benchmark comprises 150 paired cases.
All systems are evaluated using the same source videos, reference images, editing prompts, and temporal sampling protocol.
Figure~\ref{fig:editing-gsb} reports the aggregate GSB outcomes for overall quality, video quality, temporal consistency, and semantic adherence.
The corresponding mean consistency scores are 3.56 for \ourse, 2.59 for Decart-Lucy2.5, and 1.67 for XMax-X2.0.
Across the reported criteria, \ourse receives a higher preference share than both commercial baselines, with the largest observed advantages in overall quality and temporal consistency.




\begin{figure*}[!t]
\centering
\begin{subfigure}[t]{0.49\textwidth}
\centering
\includegraphics[width=\linewidth]{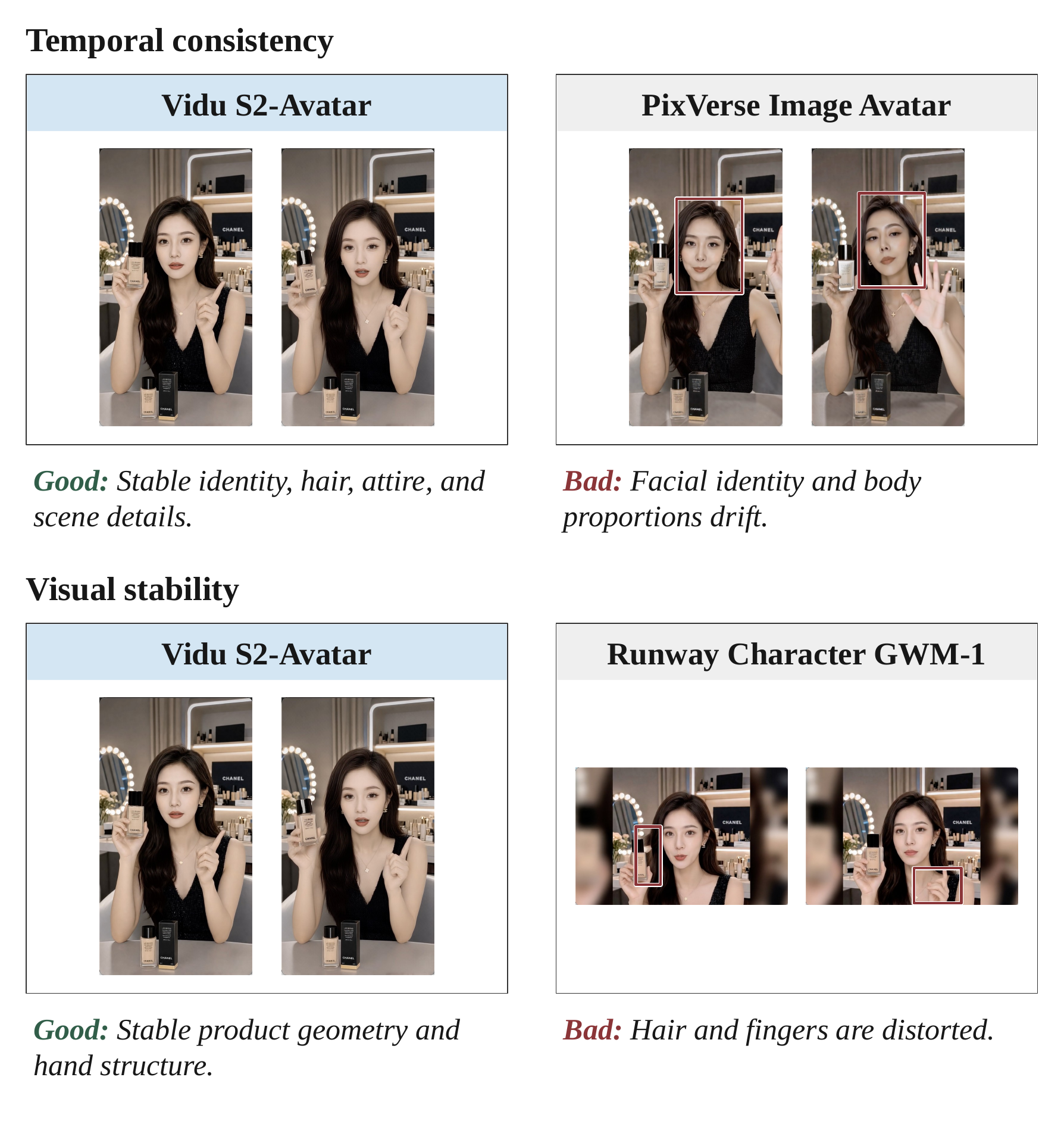}
\caption{Female-character case with cosmetics and hand interaction.}
\end{subfigure}\hfill
\begin{subfigure}[t]{0.49\textwidth}
\centering
\includegraphics[width=\linewidth]{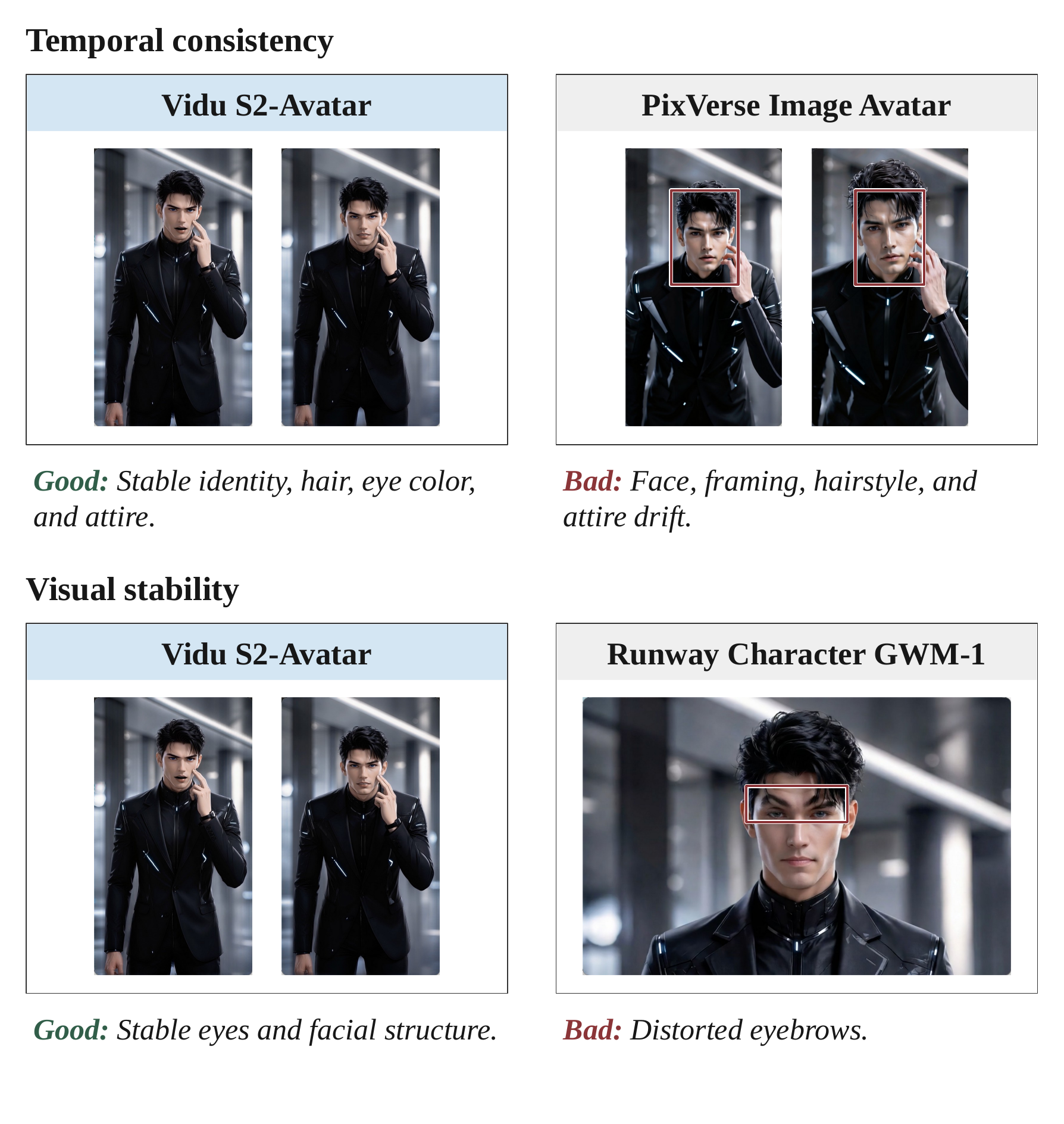}
\caption{Male-character case with facial and attire consistency.}
\end{subfigure}
\caption{Qualitative comparisons for streaming digital character generation. \oursa preserves identity, appearance attributes, fine-grained geometry, and temporal structure, while the closed-source baselines exhibit facial drift, body-proportion changes, hairstyle or attire changes, and local distortions in hair, eyebrows, fingers, or hand structure.}
\label{fig:avatar-qualitative}
\end{figure*}

\begin{figure*}[!t]
\centering
\begin{subfigure}[t]{0.49\textwidth}
\centering
\includegraphics[width=\linewidth]{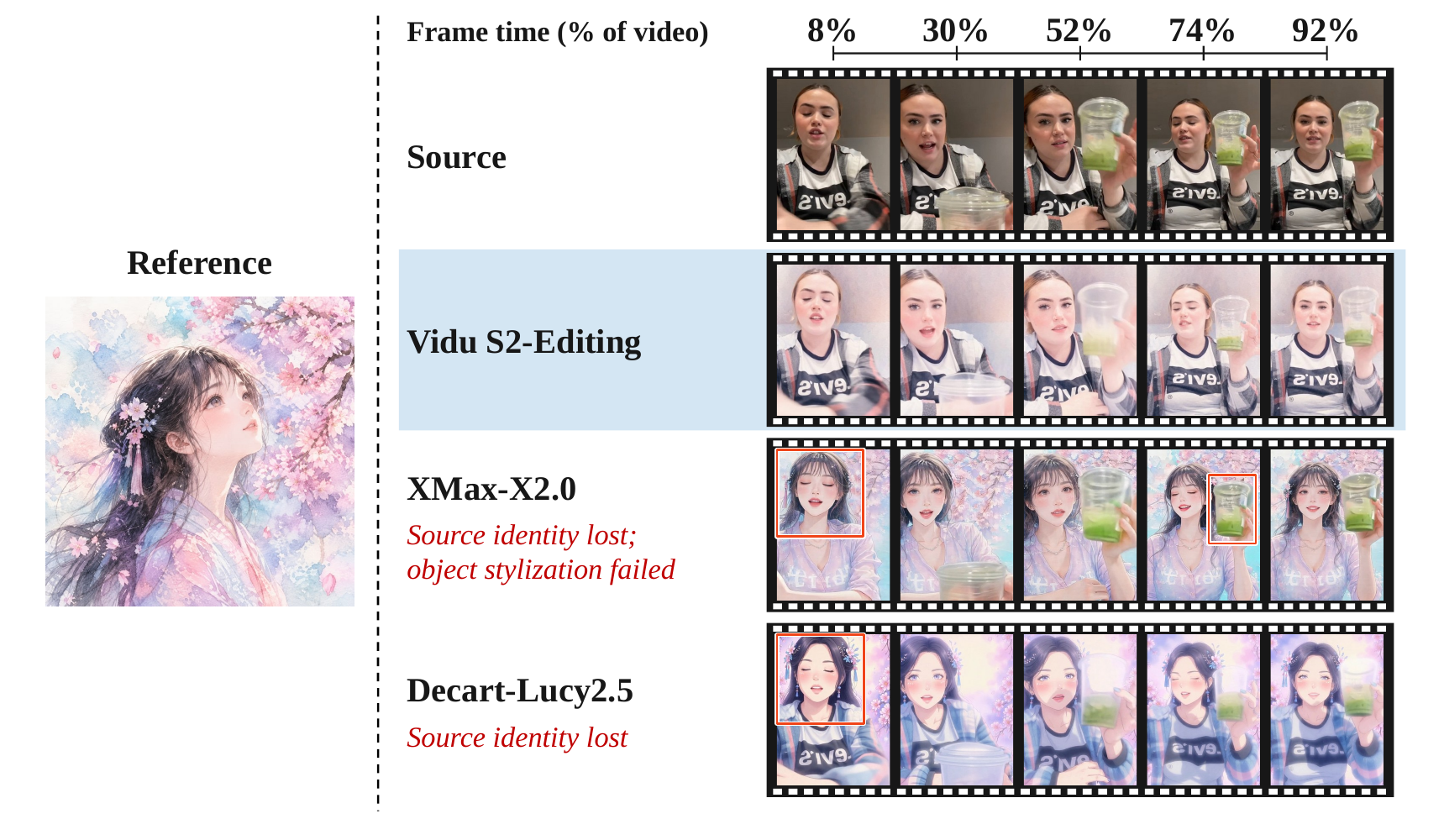}
\caption{Watercolor style transfer.}
\end{subfigure}\hfill
\begin{subfigure}[t]{0.49\textwidth}
\centering
\includegraphics[width=\linewidth]{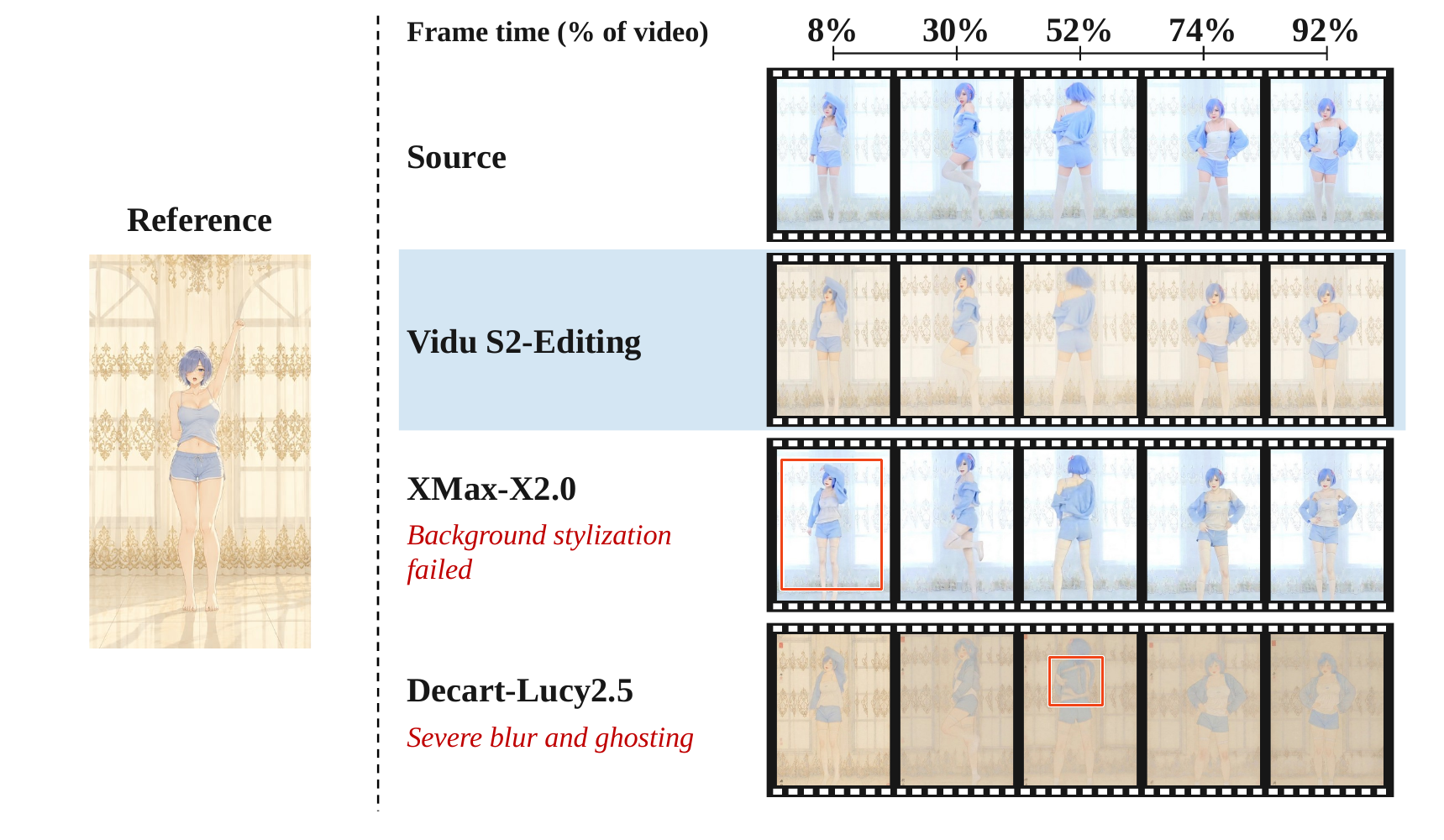}
\caption{Gongbi style transfer.}
\end{subfigure}
\caption{Style-transfer cases. \ourse changes the brushwork and palette while preserving facial layout, subject silhouette, pose, and temporal texture attachment across the stream.}
\label{fig:case-style}
\end{figure*}
\begin{figure*}[!t]
\centering
\begin{subfigure}[t]{0.49\textwidth}
\centering
\includegraphics[width=\linewidth]{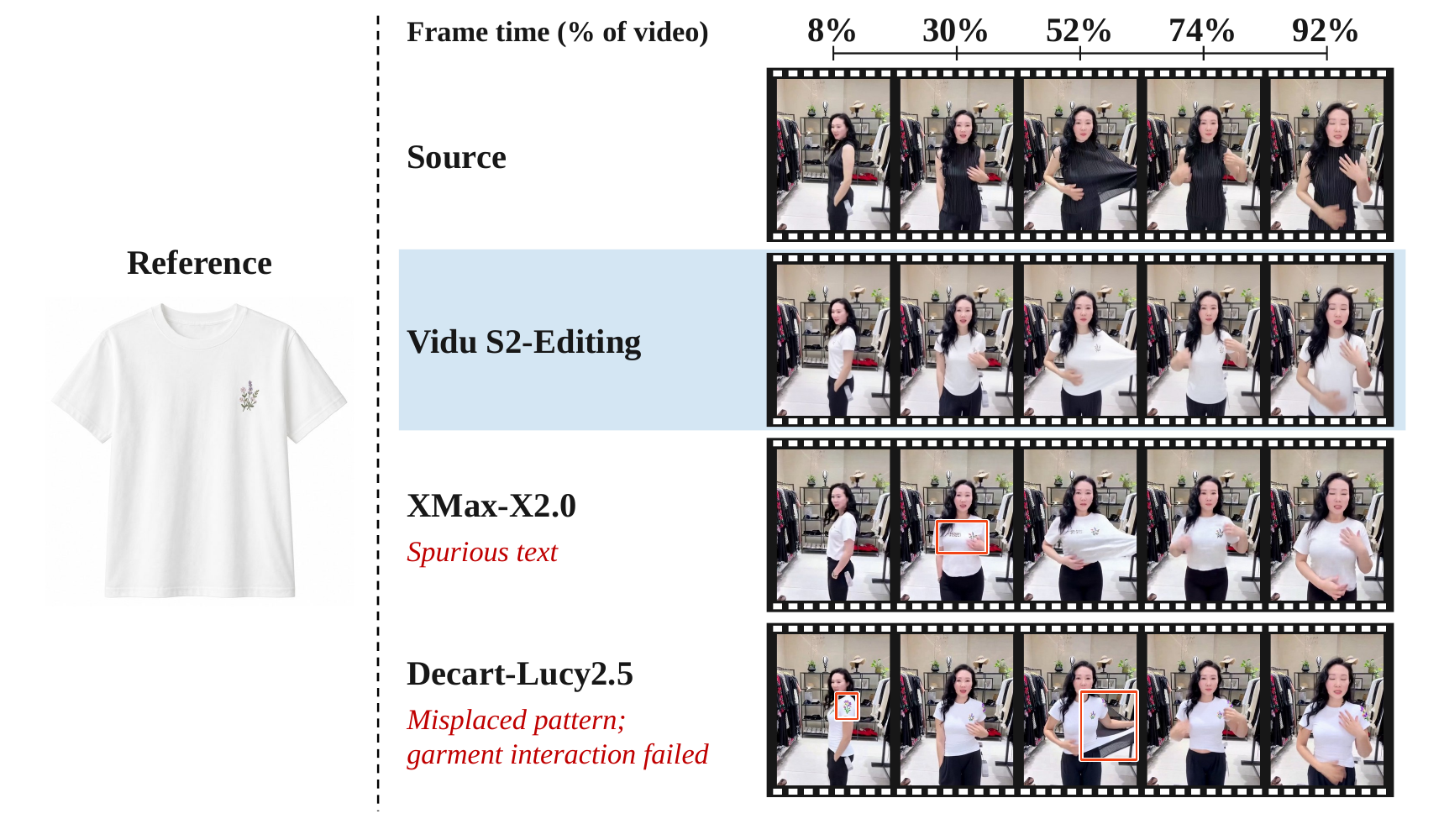}
\caption{White-shirt virtual try-on.}
\end{subfigure}\hfill
\begin{subfigure}[t]{0.49\textwidth}
\centering
\includegraphics[width=\linewidth]{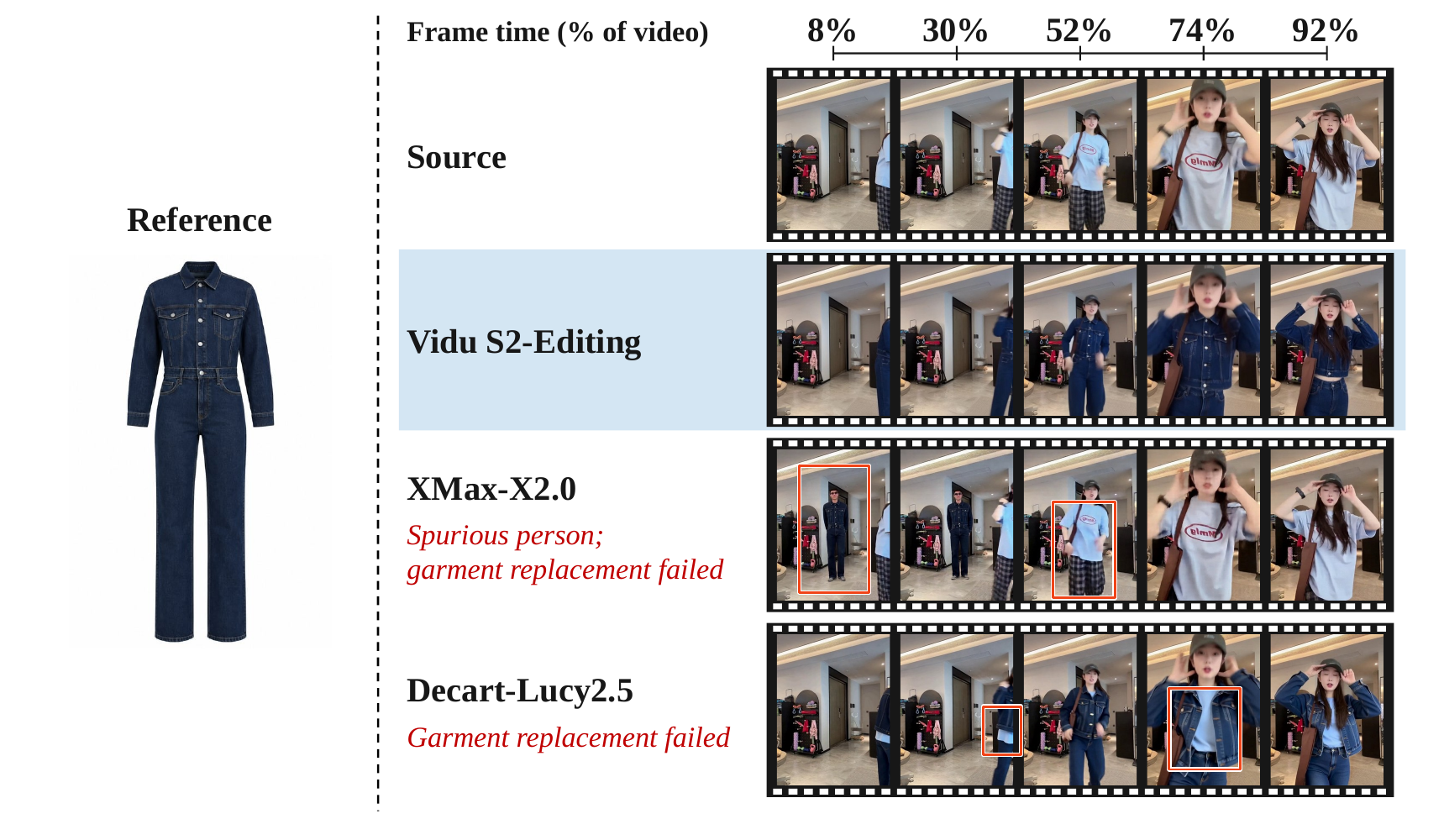}
\caption{Denim virtual try-on.}
\end{subfigure}
\caption{Virtual try-on cases. \ourse transfers the target garments while preserving body motion, garment boundaries, material texture, and hand--cloth occlusion relationships.}
\label{fig:case-tryon}
\end{figure*}

\subsection{Qualitative Case Studies}
\label{sec:exp-qualitative}

\paragraph{Vidu S2-Avatar.}

The two qualitative comparisons in Figure~\ref{fig:avatar-qualitative} examine temporal consistency and visual stability under fixed identity, scene, and motion conditions. In the female-character example, \oursa preserves the facial identity, hairstyle, facial proportions, product geometry, and hand structure across the sampled frames. PixVerse Image Avatar exhibits visible facial-identity and body-proportion drift, while Runway Character GWM-1 introduces distortions in the hair and fingers. The male-character example leads to the same conclusion: \oursa maintains the identity, hairstyle, eye color, attire, and facial structure, whereas PixVerse changes the framing, hairstyle, and overall appearance and Runway distorts the eyebrows. These cases indicate that \oursa preserves both semantic identity and fine-grained geometry over time, including small objects and articulated hand or facial details that are particularly sensitive to temporal inconsistency. The qualitative evidence get the same results.

\begin{figure*}[!t]
\centering
\begin{subfigure}[t]{0.49\textwidth}
\centering
\includegraphics[width=\linewidth]{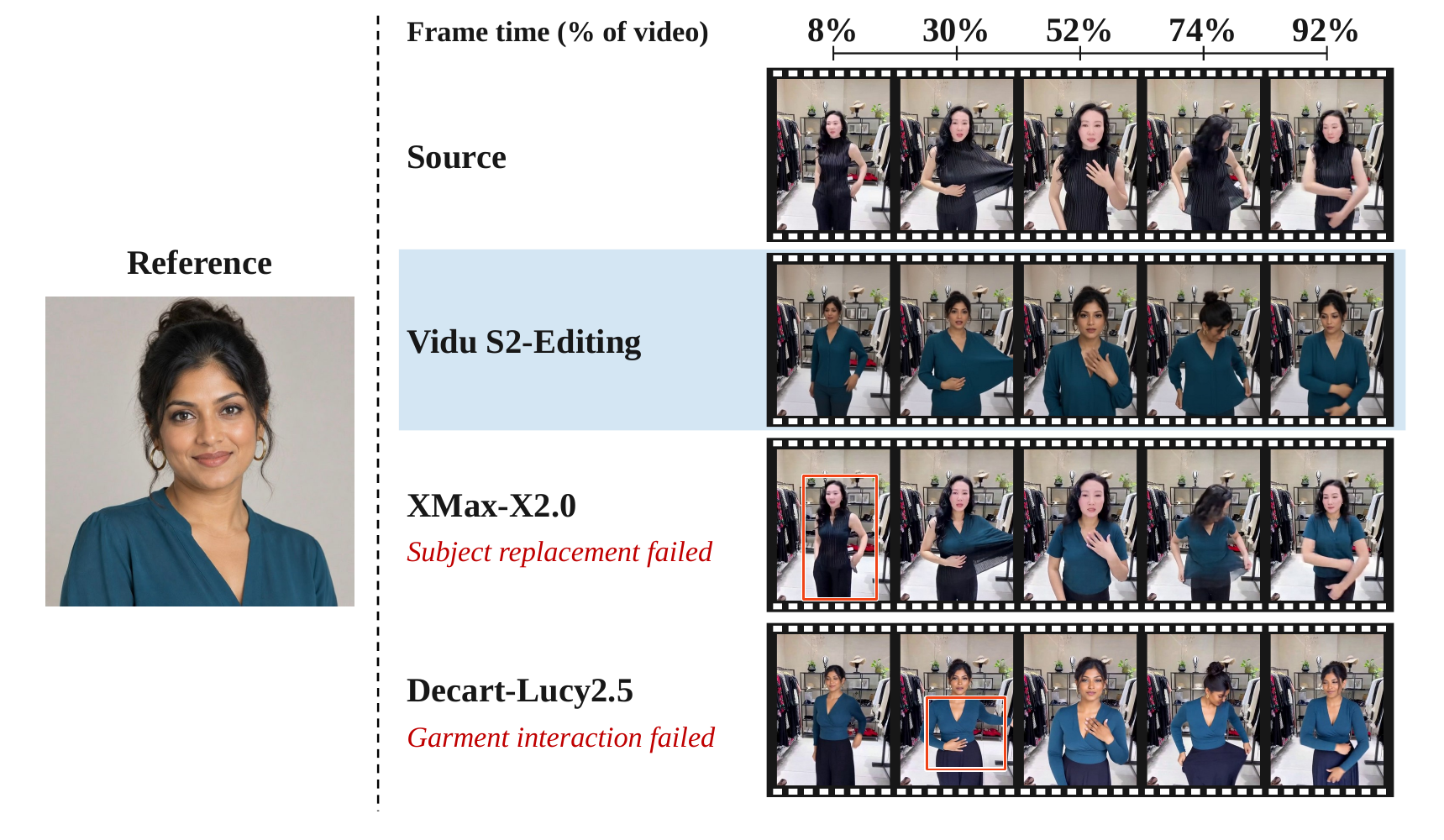}
\caption{Character subject replacement.}
\end{subfigure}\hfill
\begin{subfigure}[t]{0.49\textwidth}
\centering
\includegraphics[width=\linewidth]{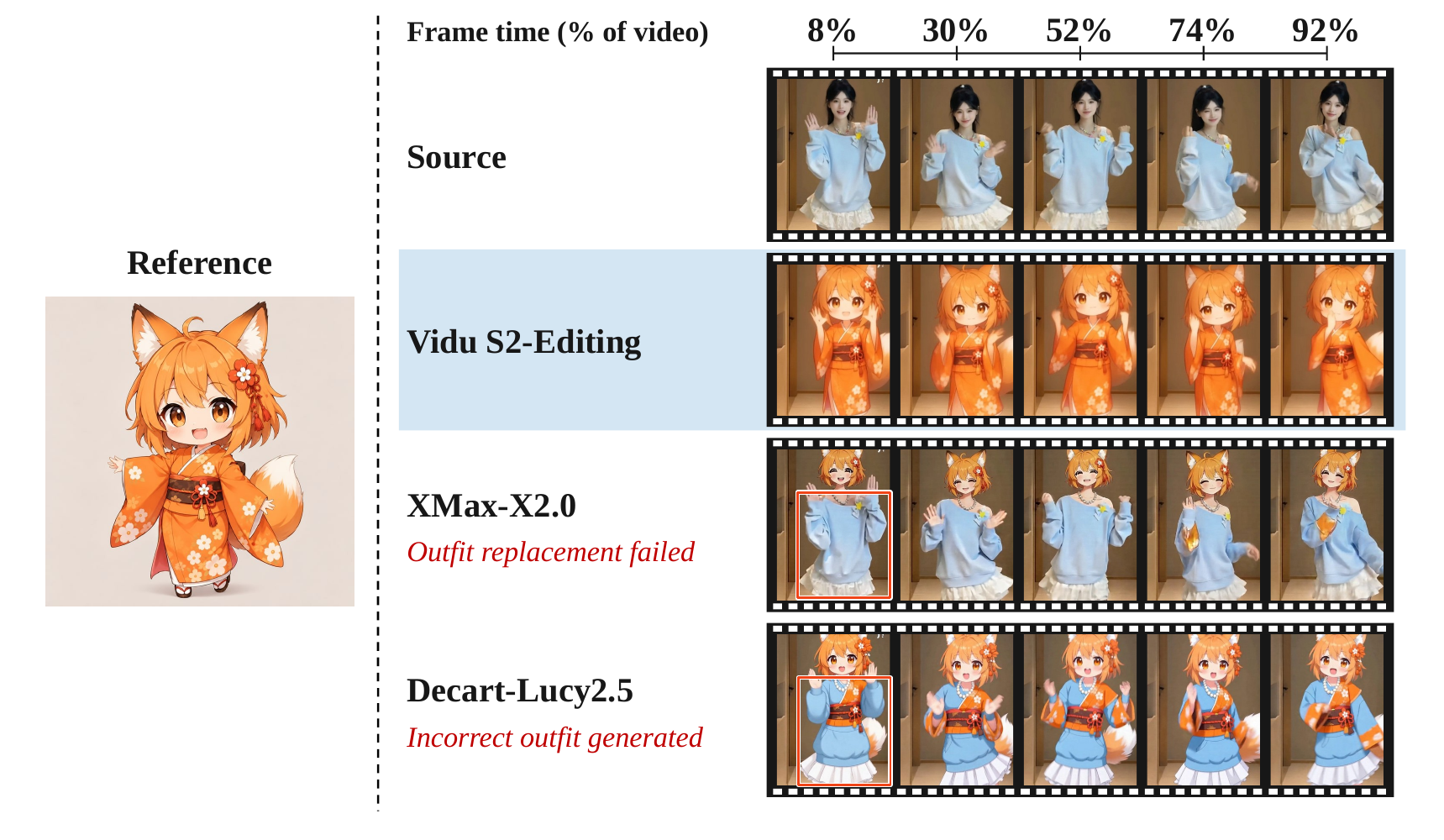}
\caption{Kimono subject replacement.}
\end{subfigure}
\caption{Subject-replacement cases. \ourse transfers the reference identity while preserving the source pose, camera trajectory, scene geometry, and coherent boundaries around hair and limbs.}
\label{fig:case-subject}
\end{figure*}

\begin{figure*}[!t]
\centering
\begin{subfigure}[t]{0.49\textwidth}
\centering
\includegraphics[width=\linewidth]{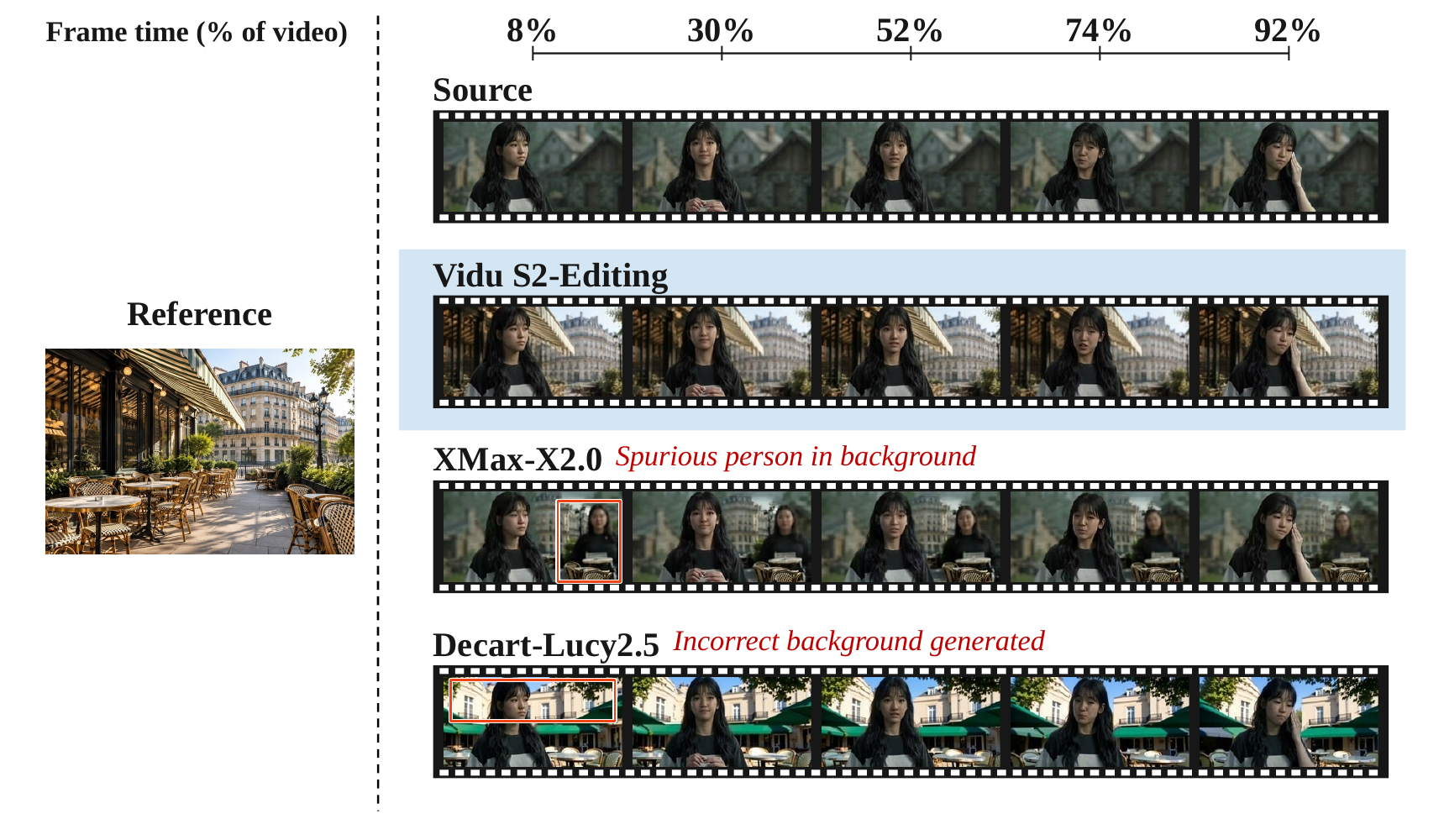}
\caption{Paris background replacement.}
\end{subfigure}\hfill
\begin{subfigure}[t]{0.49\textwidth}
\centering
\includegraphics[width=\linewidth]{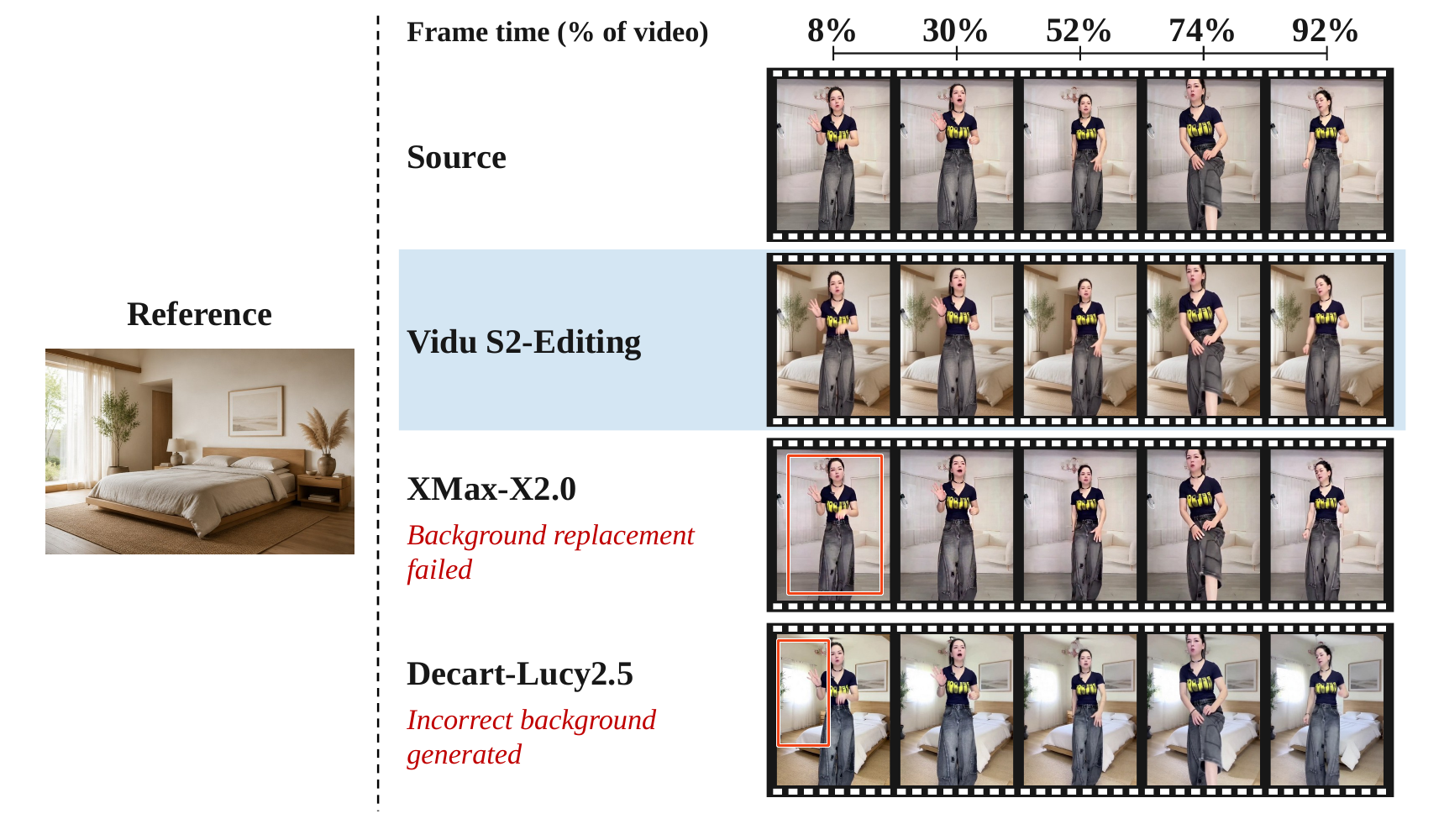}
\caption{Bedroom background replacement.}
\end{subfigure}
\caption{Background-replacement cases. \ourse changes the target environment while preserving foreground pose, camera motion, scene geometry, and closed boundaries around fine contours over time.}
\vspace{-1em}
\label{fig:case-background}
\end{figure*}

\paragraph{Vidu S2-Editing.}
We compare four editing tasks using the cases in Figures~\ref{fig:case-style}--\ref{fig:case-background}: style transfer, virtual try-on, subject replacement, and background replacement.
(1) \textit{Style Transfer.}
Figure~\ref{fig:case-style} compares watercolor and gongbi stylization.
In the watercolor case, \ourse applies the reference palette and rendering style while retaining the source subject's facial appearance and interaction with the cup.
XMax-X2.0 changes the subject's identity but leaves the cup with a photographic appearance, failing to stylize the scene consistently.
Decart-Lucy2.5 also changes the source facial identity.
In the gongbi case, XMax-X2.0 fails to stylize the background, while Decart-Lucy2.5 produces severe blur and ghosting around the subject.
\ourse applies the reference style to both the subject and background while retaining clearer subject contours in the displayed frames.
(2) \textit{Virtual Try-On.}
Figure~\ref{fig:case-tryon} compares transfer of a white shirt and a denim jumpsuit.
In the white-shirt case, \ourse reproduces the reference garment and floral detail while retaining the source hand--garment interaction.
XMax-X2.0 introduces spurious text on the chest.
Decart-Lucy2.5 places the floral pattern on the shoulder and fails to preserve the shirt deformation during the pulling action highlighted in the figure.
In the denim case, XMax-X2.0 inserts an extra person wearing denim while leaving the original subject in the source outfit, failing to replace the intended subject's clothing.
Decart-Lucy2.5 generates a denim jacket over the source shirt instead of the reference jumpsuit.
\ourse transfers the jumpsuit across the displayed poses without introducing an additional person.
(3) \textit{Subject Replacement.}
Figure~\ref{fig:case-subject} compares replacement with a reference woman and a cartoon character wearing an orange kimono.
In the human case, \ourse transfers the reference appearance while retaining the source poses and shirt-pulling interaction.
XMax-X2.0 fails to replace the source subject in the highlighted frame and subsequently changes the clothing without consistently transferring the reference identity.
Decart-Lucy2.5 changes the subject's appearance but fails to reproduce the source shirt-pulling interaction in the highlighted second frame.
In the cartoon case, XMax-X2.0 changes the head while retaining the source blue outfit, and Decart-Lucy2.5 generates an incorrect outfit that combines blue source clothing with orange kimono elements.
\ourse transfers both the character appearance and the reference kimono across the displayed frames.
(4) \textit{Background Replacement.}
Figure~\ref{fig:case-background} compares replacement with a Paris cafe scene and a bedroom.
\ourse reproduces the reference environments while retaining the foreground subject and source actions.
In the Paris case, XMax-X2.0 introduces an extra person and retains the original setting instead of replacing it with the reference scene.
Decart-Lucy2.5 generates a different cafe layout whose architecture and awnings do not match the reference.
In the bedroom case, XMax-X2.0 leaves the original room unchanged, failing to replace the background.
Decart-Lucy2.5 generates a bedroom, but its window, bed, and lighting arrangement differ from the reference.
These cases show failures in both background replacement, whereas \ourse closely reproduces the requested scene layout in the displayed frames.

\begin{figure*}[h!]
\centering
\includegraphics[width=0.98\textwidth]{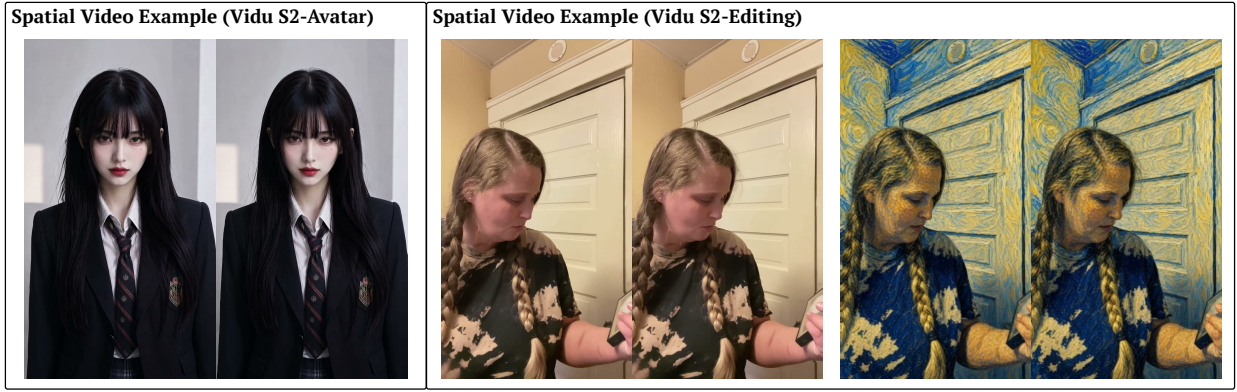}
\caption{Qualitative results of \ours spatial video generation. Each example shows the generated left- and right-eye views.}
\label{fig:spatial-video-qualitative}
\end{figure*}

\paragraph{Vidu S2 Spatial Video Generation.}
Figure~\ref{fig:spatial-video-qualitative} presents representative spatial-video results produced by \ours. Across the examples, foreground subjects are clearly separated from the background, creating coherent scene depth and a stronger sense of immersion.

\section{Conclusion}
We present \ours, which comprises \oursa, a real-time interactive digital-character model, and \ourse, a real-time video editing model, and we further explore the feasibility of real-time spatial video generation.
Compared with \previous, \oursa supports real-time 720p video generation, dynamic references that can be updated at any moment, and stronger instruction following, such as dancing.
\ourse edits a video stream in real time, covering style transfer, virtual try-on, character replacement, and background replacement.
Experiments show that \ours outperforms all baselines.


\bibliographystyle{unsrt}
\bibliography{main}

\end{document}